\documentclass{article}
\usepackage{graphicx, subcaption, siunitx, float, hyperref, cleveref, booktabs}
\usepackage[utf8]{inputenc}

\hypersetup{
    colorlinks=true,
    linkcolor=black,
    citecolor=black,
    urlcolor=blue,
}

\usepackage{natbib}
\setcitestyle{authoryear,open={(},close={)}}

\newcommand{\changeurlcolor}[1]{\hypersetup{urlcolor=#1}} 

\nocite{*}

\usepackage[accepted]{icml2020}

\icmltitlerunning{Leveraging Procedural Generation to Benchmark Reinforcement Learning}

\begin{document}

\twocolumn[
\icmltitle{Leveraging Procedural Generation to Benchmark Reinforcement Learning}

\icmlsetsymbol{equal}{*}

\begin{icmlauthorlist}
\icmlauthor{Karl Cobbe}{op}
\icmlauthor{Christopher Hesse}{op}
\icmlauthor{Jacob Hilton}{op}
\icmlauthor{John Schulman}{op}
\end{icmlauthorlist}

\icmlaffiliation{op}{OpenAI, San Francisco, CA, USA}

\icmlcorrespondingauthor{Karl Cobbe}{karl@openai.com}

\icmlkeywords{Machine Learning, Reinforcement Learning, Transfer Learning, Generalization, Procedural Generation, ICML}

\vskip 0.3in
]

\printAffiliationsAndNotice{}

\begin{abstract}
We introduce Procgen Benchmark, a suite of 16 procedurally generated game-like environments designed to benchmark both sample efficiency and generalization in reinforcement learning. We believe that the community will benefit from increased access to high quality training environments, and we provide detailed experimental protocols for using this benchmark. We empirically demonstrate that diverse environment distributions are essential to adequately train and evaluate RL agents, thereby motivating the extensive use of procedural content generation. We then use this benchmark to investigate the effects of scaling model size, finding that larger models significantly improve both sample efficiency and generalization.

\end{abstract}

\section{Introduction}

Generalization remains one of the most fundamental challenges in deep reinforcement learning. In several recent studies \citep{study_overfitting, coinrun, illuminating_gen, obstacle}, agents exhibit the capacity to overfit to remarkably large training sets. This evidence raises the possibility that overfitting pervades classic benchmarks like the Arcade Learning Environment (ALE) \citep{ale}, which has long served as a gold standard in RL. While the diversity between games in the ALE is one of the benchmark's greatest strengths, the low emphasis on generalization presents a significant drawback. Previous work has sought to alleviate overfitting in the ALE by introducing sticky actions \citep{machado18arcade} or by embedding natural videos as backgrounds \citep{natural_ale}, but these methods only superficially address the underlying problem --- that agents perpetually encounter near-identical states. For each game the question must be asked: are agents robustly learning a relevant skill, or are they approximately memorizing specific trajectories?

There have been several investigations of generalization in RL \citep{gen_dqn, assess_gen_rl, dissect_overfitting, lee2019simple}, but progress has largely proved elusive. Arguably one of the principal setbacks has been the lack of environments well-suited to measure generalization. While previously mentioned studies \citep{study_overfitting, coinrun, illuminating_gen, obstacle} reveal intriguing trends, it is hard to draw general conclusions from so few environments.

We seek the best of both worlds: a benchmark with overall diversity comparable to the ALE, comprised of environments that fundamentally require generalization. We have created Procgen Benchmark to fulfill this need. This benchmark is ideal for evaluating generalization, as distinct training and test sets can be generated for each environment. This benchmark is also well-suited to evaluate sample efficiency, as all environments pose diverse and compelling challenges for RL agents. The environments' intrinsic diversity demands that agents learn robust policies; overfitting to narrow regions in state space will not suffice. Put differently, the ability to generalize becomes an integral component of success when agents are faced with ever-changing levels. All environments are open-source and can be found at \href{https://github.com/openai/procgen}{https://github.com/openai/procgen}.

\begin{figure}
\begin{subfigure}{.48 \textwidth}
\centering
\includegraphics[width=.24 \textwidth]{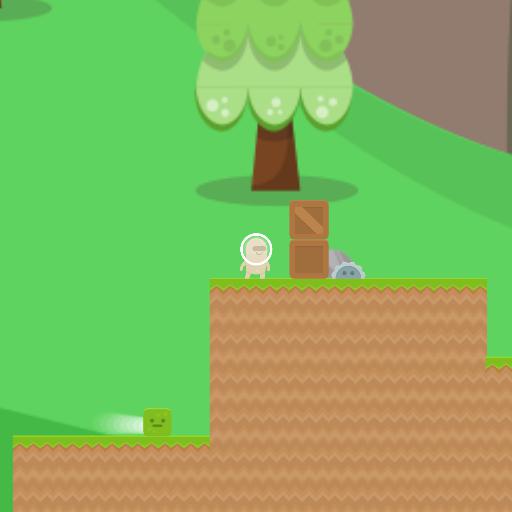}
\includegraphics[width=.24 \textwidth]{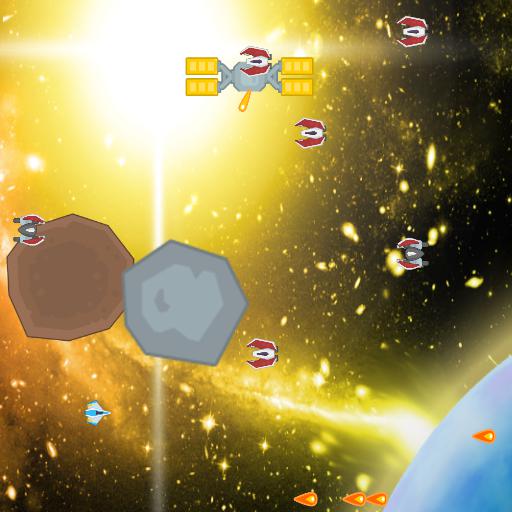}
\includegraphics[width=.24 \textwidth]{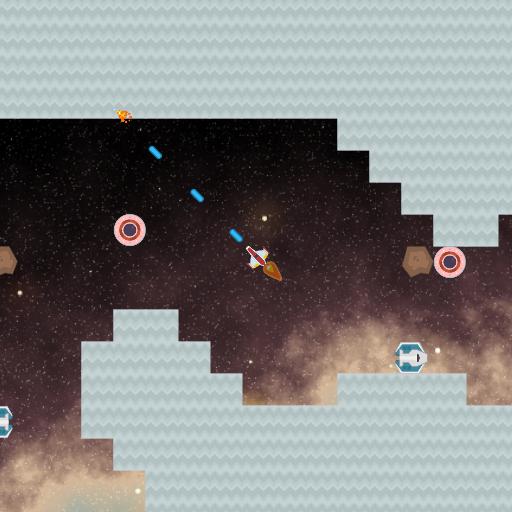}
\includegraphics[width=.24 \textwidth]{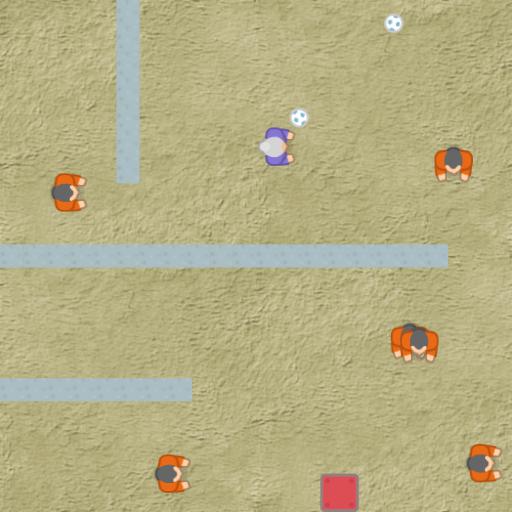}
\end{subfigure}
\begin{subfigure}{.48 \textwidth}
\centering
\vspace*{.75mm}
\includegraphics[width=.24 \textwidth]{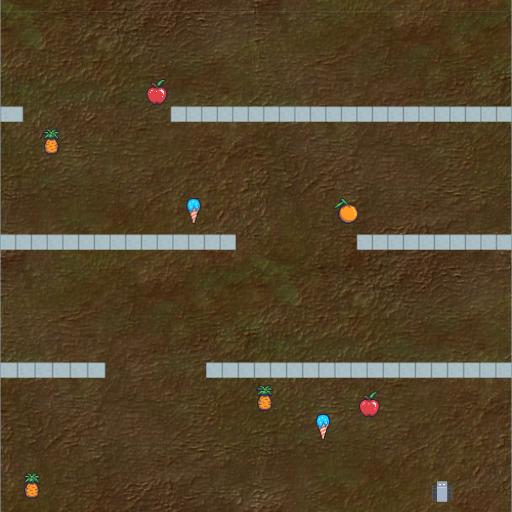}
\includegraphics[width=.24 \textwidth]{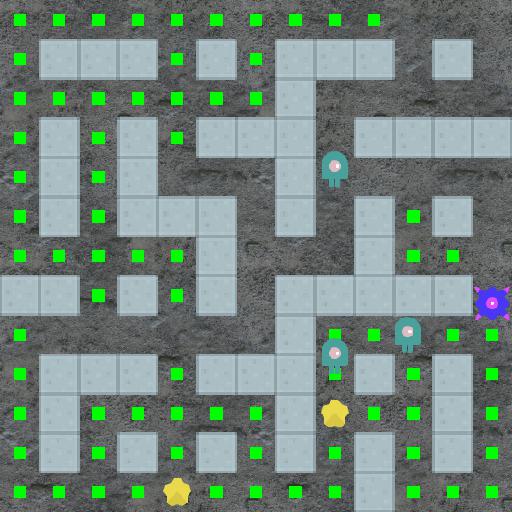}
\includegraphics[width=.24 \textwidth]{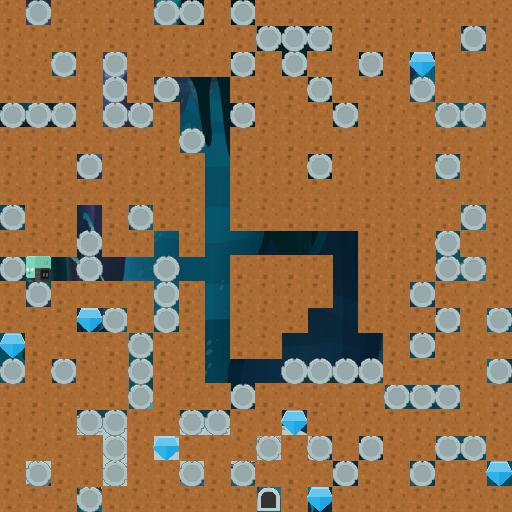}
\includegraphics[width=.24 \textwidth]{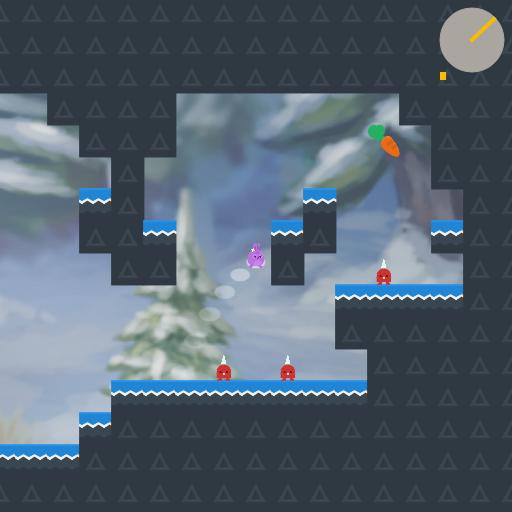}
\end{subfigure}
\begin{subfigure}{.48 \textwidth}
\centering
\vspace*{.75mm}
\includegraphics[width=.24 \textwidth]{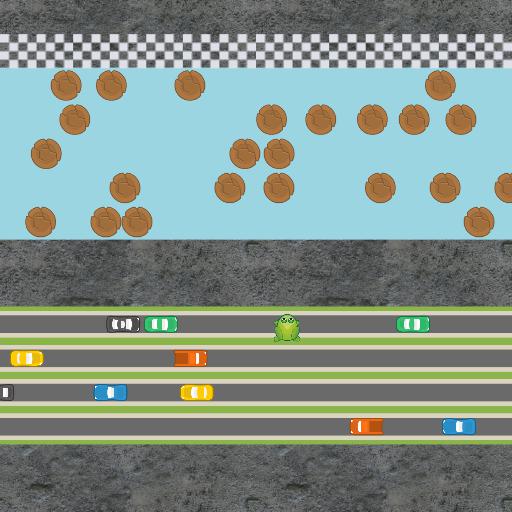}
\includegraphics[width=.24 \textwidth]{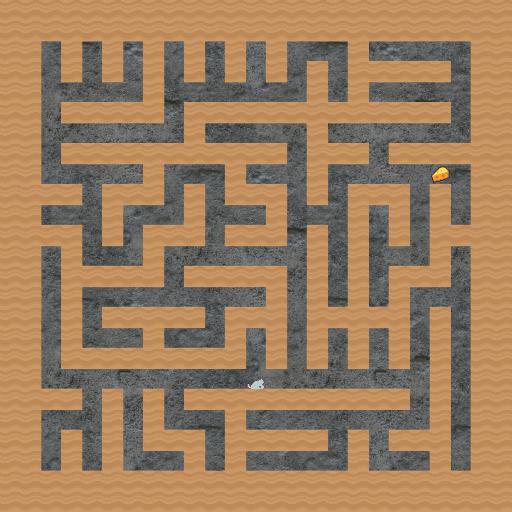}
\includegraphics[width=.24 \textwidth]{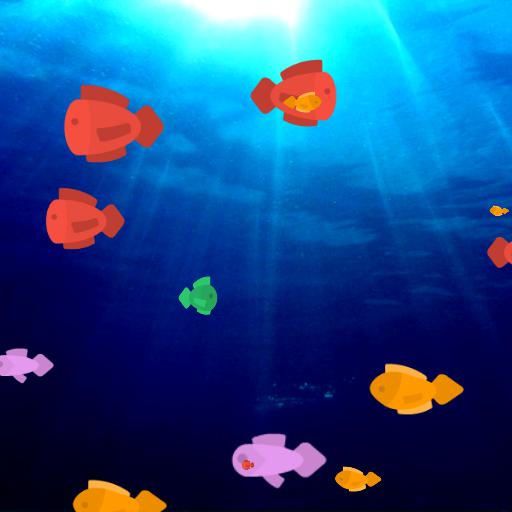}
\includegraphics[width=.24 \textwidth]{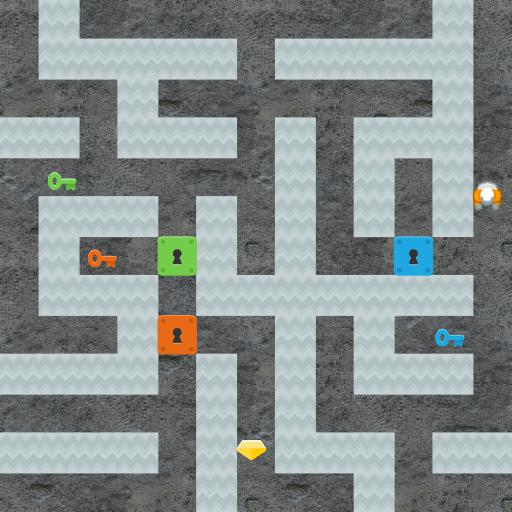}
\end{subfigure}
\begin{subfigure}{.48 \textwidth}
\centering
\vspace*{.75mm}
\includegraphics[width=.24 \textwidth]{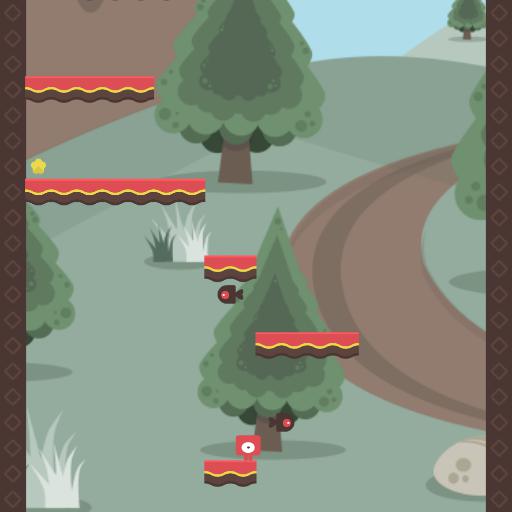}
\includegraphics[width=.24 \textwidth]{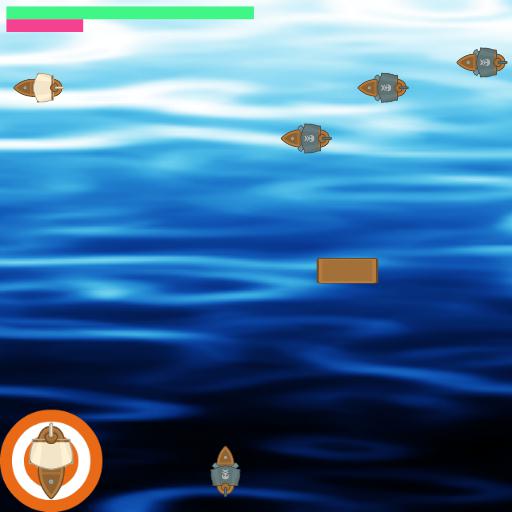}
\includegraphics[width=.24 \textwidth]{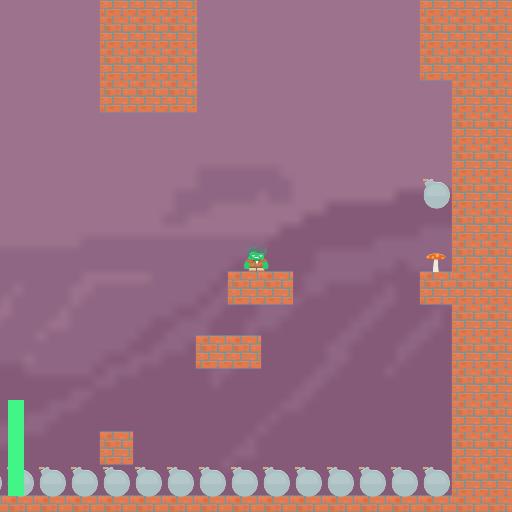}
\includegraphics[width=.24 \textwidth]{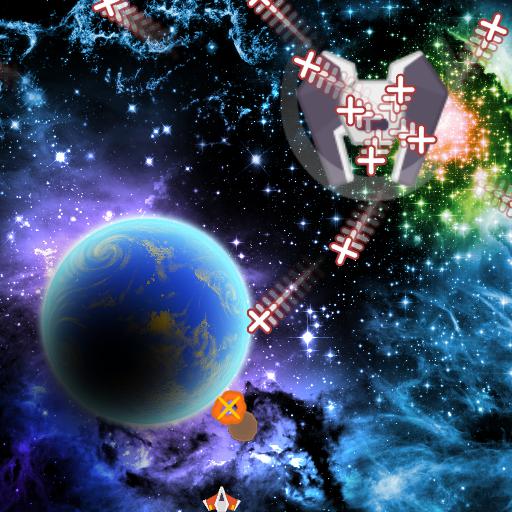}
\end{subfigure}
\caption{Screenshots from each game in Procgen Benchmark.}
\label{fig:all_screenshots}
\end{figure}

\section{Procgen Benchmark}

Procgen Benchmark consists of 16 unique environments designed to measure both sample efficiency and generalization in reinforcement learning. These environments greatly benefit from the use of procedural content generation, the algorithmic creation of a near-infinite supply of highly randomized content. In these environments, employing procedural generation is far more effective than relying on fixed, human-designed content.

Procedural generation logic governs the level layout \citep{johnson2010cellular}, the selection of game assets, the location and spawn times of entities, and other game-specific details. To master any one of these environments, agents must learn a policy that is robust across all axes of variation. Learning such a policy is both more challenging and more relevant than overfitting to a handful of fixed levels. Screenshots from each environment are shown in \Cref{fig:all_screenshots}. We note that the state transition function is deterministic in all environments.\footnote{Although the Chaser environment is deterministic, the enemy AI will make pseudorandom decisions conditioned on the level seed.}

\subsection{Environment Desiderata}

We designed all environments to satisfy the following criteria. \\

\textbf{High Diversity:} Procedural generation logic is given maximal freedom, subject to basic design constraints. The diversity in the resulting level distributions presents agents with meaningful generalization challenges. \\

\textbf{Fast Evaluation:} Environment difficulty is calibrated such that baseline agents make significant progress training over 200M timesteps. Moreover, the environments are optimized to perform thousands of steps per second on a single CPU core, including the time required to render observations. This enables a fast experimental pipeline. \\

\textbf{Tunable Difficulty:} All environments support two well-calibrated difficulty settings: easy and hard. This difficulty refers to the level distribution and not to individual levels; in both settings, the difficulty of individual levels has high variance. Unless otherwise specified, we report results using the hard difficulty setting. We make the easy difficulty setting available for those with limited access to compute power, as it reduces the resources required to train agents by roughly a factor of 8. \\

\textbf{Level Solvability:} The procedural generation in each environment strives to make all levels solvable, but this is not strictly guaranteed. For each environment, greater than 99\% of levels are believed to be solvable. \\

\textbf{Emphasis on Visual Recognition and Motor Control:} In keeping with precedent, environments mimic the style of many Atari and Gym Retro \citep{gymretro} games. Performing well primarily depends on identifying critical assets in the observation space and enacting appropriate low level motor responses. \\

\textbf{Shared Action and Observation Space:} To support a unified training pipeline, all environments use a discrete 15 dimensional action space and produce $64\times64\times3$ RGB observations. Some environments include no-op actions to accommodate the shared action space. \\

\textbf{Tunable Dependence on Exploration:} These environments were designed to be tractable for baseline RL agents without the need for custom exploratory rewards. However, many of these environments can be made into more challenging exploration tasks if desired. See \Cref{appendix:exploration} for a discussion on evaluating exploration capability. \\

\textbf{Tunable Dependence on Memory:} These environments were designed to require minimal use of memory, in order to isolate the challenges in RL. However, several environments include variants that do test the use of memory, as we discuss in \Cref{appendix:memory}. \\

By satisfying these requirements, we believe Procgen Benchmark will be a valuable tool in RL research. Descriptions of each specific environment can be found in \Cref{appendix:env_disc}.

\begin{figure*}
\centering
\includegraphics[width=\textwidth]{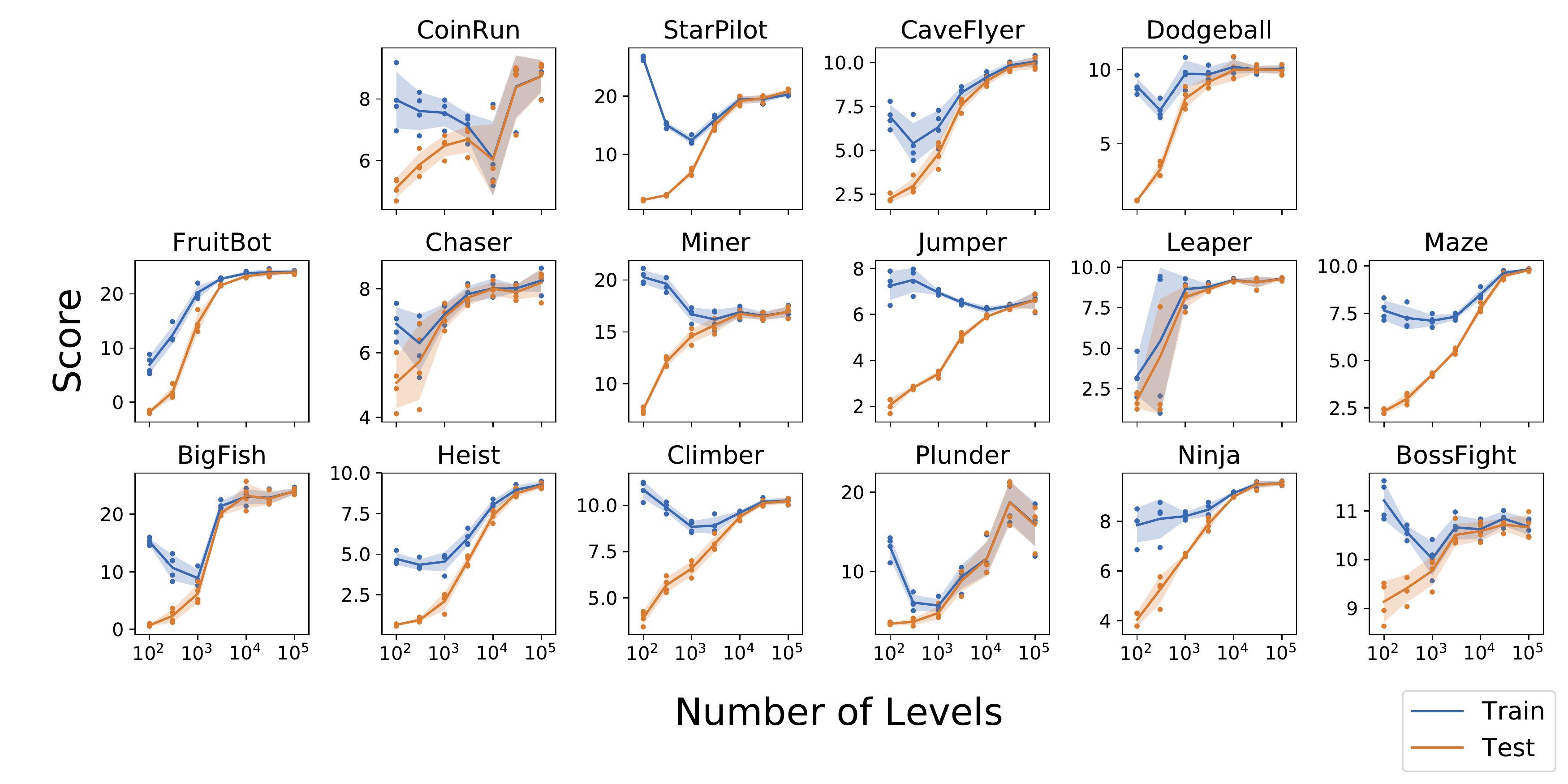}
\caption{Generalization performance in each environment as a function of training set size. We report the mean raw episodic return, where each episode includes a single level. The mean and standard deviation is shown across 4 seeds.}
\label{fig:gen_all}
\end{figure*}

\subsection{Experimental Protocols} \label{sec:protocols}

By default, we train agents using Proximal Policy Optimization \citep{ppo} for 200M timesteps. While this timestep choice is arbitrary, it follows the precedent set by the ALE. It is also experimentally convenient: training for 200M timesteps with PPO on a single Procgen environment requires approximately 24 GPU-hrs and 60 CPU-hrs. We consider this a reasonable and practical computational cost. To further reduce training time at the cost of experimental complexity, environments can be set to the easy difficulty. We recommend training easy difficulty environments for 25M timesteps, which requires approximately 3 GPU-hrs with our implementation of PPO.

When evaluating sample efficiency, we train and test agents on the full distribution of levels in each environment. When evaluating generalization, we train on a finite set of levels and we test on the full distribution of levels. Unless otherwise specified, we use a training set of 500 levels to evaluate generalization in each environment. For easy difficulty environments, we recommend using training sets of 200 levels. We report results on easy difficulty environments in \Cref{appendix:easy}.

When it is necessary to report a single score across Procgen Benchmark, we calculate the mean normalized return. For each environment, we define the normalized return to be $R_{norm} = (R - R_{min}) / (R_{max} - R_{min})$, where $R$ is the raw expected return and $R_{min}$ and $R_{max}$ are constants chosen to approximately bound $R$. Under this definition, the normalized return will almost always fall between 0 and 1. We use the mean normalized return as it provides a better signal than the median, and since there is no need to be robust to outliers. We designed all environments to have similar difficulties in order to prevent a small subset from dominating this signal. See \Cref{appendix:norm_constants} for a list of normalization constants and a discussion on their selection.

\subsection{Hyperparameter Selection}

In deep RL, hyperparameter tuning is often the difference between great and mediocre results. Unfortunately, this process can be costly in both time and computation. For those who are more comfortable with the existing ALE benchmark, minimal hyperparameter tuning should be required to train on Procgen environments. This is partially by design, as Procgen Benchmark heavily draws inspiration from the ALE and Gym Retro. To provide a point of comparison, we evaluate our Procgen-tuned implementation of PPO on the ALE, and we achieve competitive performance. Detailed results are shown in \Cref{appendix:atari}.

As a convenience, we choose not to use any frame stacking in Procgen experiments, as we find this only minimally impacts performance. See \Cref{appendix:frame_stack} for further discussion. By default, we train agents with the convolutional architecture found in IMPALA \citep{impala}, as we find this architecture strikes a reasonable balance between performance and compute requirements. We note that smaller architectures often struggle to train when faced with the high diversity of Procgen environments, a trend we explore further in \Cref{sec:network_size}.

\begin{figure*}
\centering
\includegraphics[width=\textwidth]{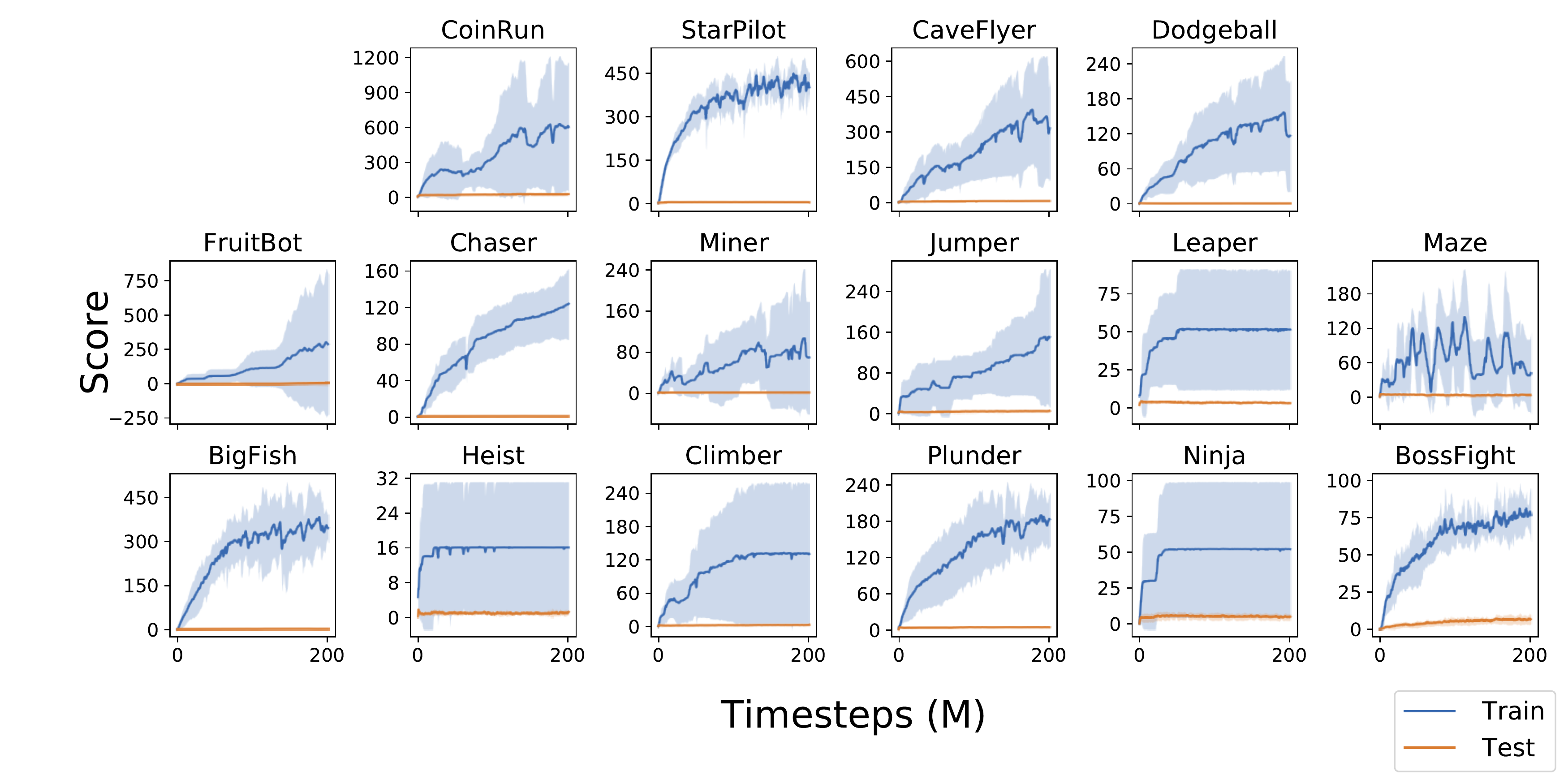}
\caption{Train and test performance when training with a deterministic sequence of levels. We report the mean raw episodic return, where each episode may include many sequential levels. The mean and standard deviation is shown across 4 seeds.}
\label{fig:seq}
\end{figure*}

\section{Generalization Experiments} \label{sec:eval_gen}

\subsection{Level Requirements} \label{sec:level_gen}

We first evaluate the impact of training set size on generalization. For each environment, we construct several training sets ranging in size from 100 to 100,000 levels. We train agents for 200M timesteps on each training set using PPO, and we measure performance on held out levels. Results are shown in \Cref{fig:gen_all}. See \Cref{appendix:hyperparameters} for a list of hyperparameters and \Cref{appendix:gen_test_curves} for test curves from each training set.

We find that agents strongly overfit to small training sets in almost all cases. To close the generalization gap, agents need access to as many as 10,000 levels. A peculiar trend emerges in many environments: past a certain threshold, training performance improves as the training set grows. This runs counter to trends found in supervised learning, where training performance commonly decreases with the size of the training set. We attribute this trend to the implicit curriculum provided by the distribution of levels. A larger training set can improve training performance if the agent learns to generalize even across levels in the training set. This effect was previously reported by \cite{coinrun}, and we now corroborate those results with a larger number of environments.

\subsection{An Ablation with Deterministic Levels}

To fully emphasize the significance of procedural generation, we conduct a simple ablation study. Instead of re-sampling a new level at the start of every episode, we train agents on a fixed sequence of levels. In each episode, the agent begins on the first level. When the agent successfully completes a level, it progresses to the next level. If the agent fails at any point, the episode terminates. With this setup, the agent can reach arbitrarily many levels, though in practice it rarely progresses beyond the $20\textsuperscript{th}$ level in any environment. This approximately mimics the training setup of the ALE. To make training more tractable in this setting, we use the easy environment difficulty.

\begin{figure*}
\centering
\begin{subfigure}{0.475 \textwidth}
\includegraphics[width=\textwidth]{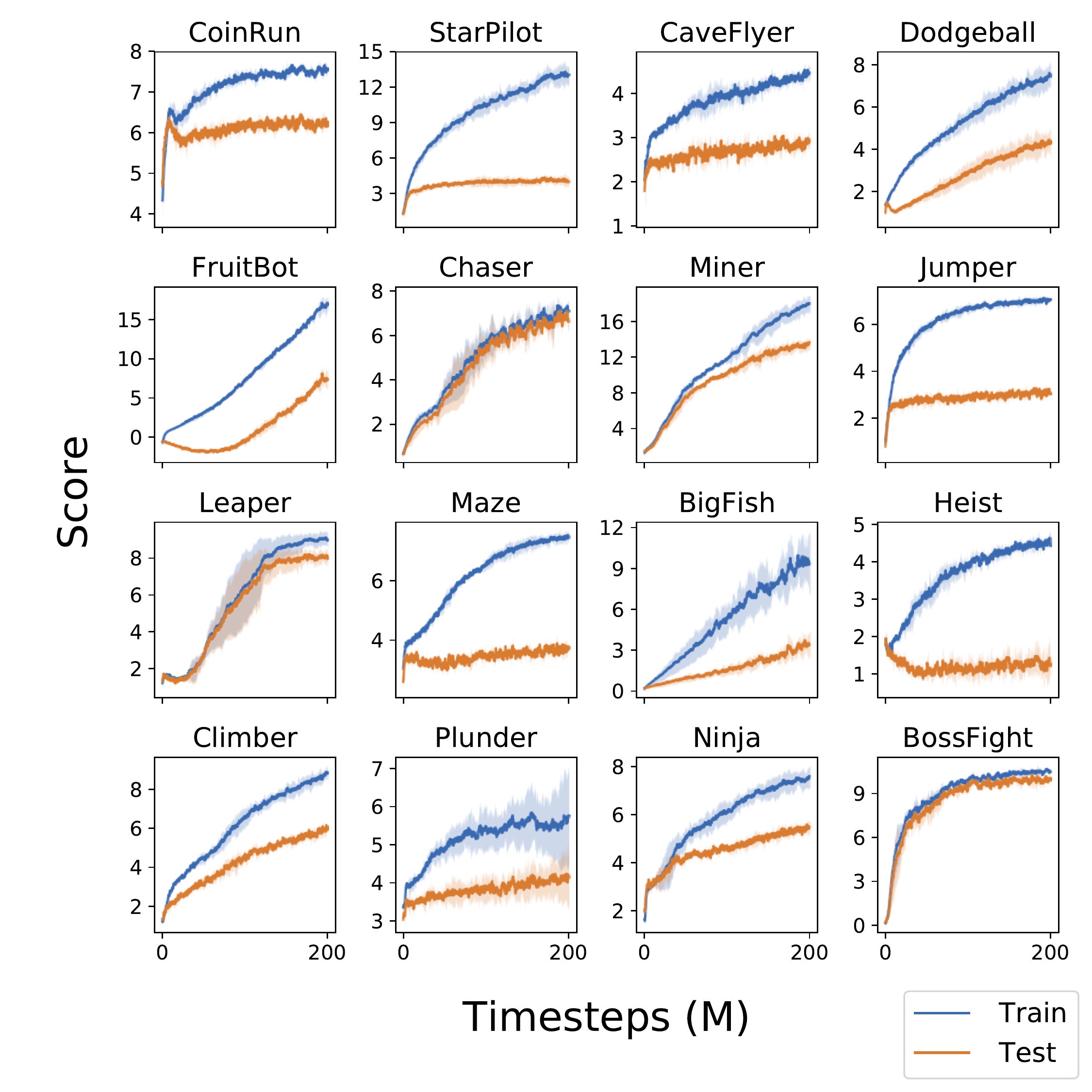}
\end{subfigure}
\hspace*{\fill}
\begin{subfigure}{0.475 \textwidth}
\includegraphics[width=\textwidth]{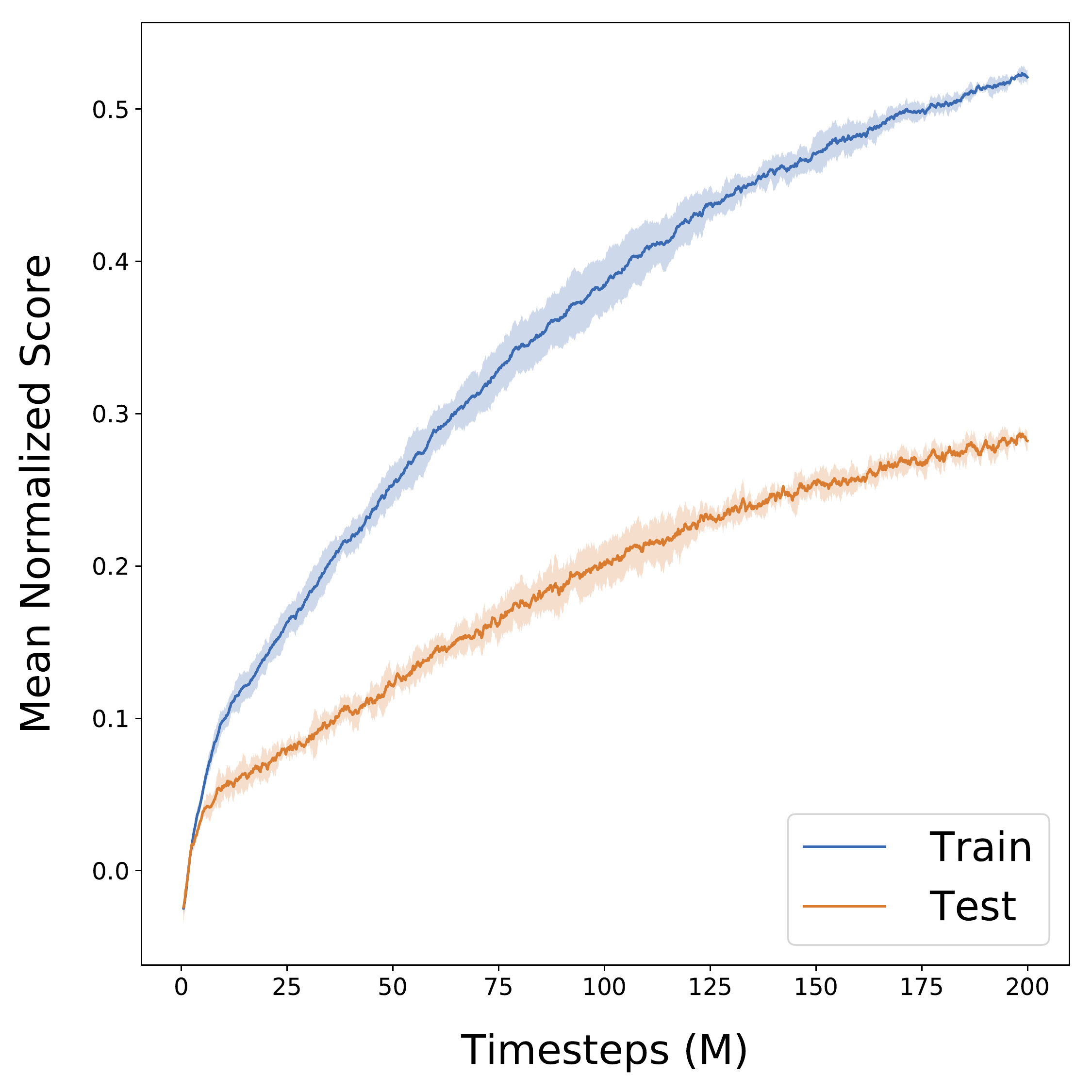}
\end{subfigure}
\caption{Generalization performance from 500 levels in each environment. The mean and standard deviation is shown across 3 seeds.}
\label{fig:gen500}
\end{figure*}

At test time, we simply remove the determinism in the level sequence, instead choosing level sequences at random. Results are shown in \Cref{fig:seq}. We find that agents become competent over the first several training levels in most environments, giving an illusion of meaningful progress. However, test performance demonstrates that the agents have in fact learned almost nothing about the underlying level distribution. We believe this vast gap between train and test performance is worth highlighting. It reveals a crucial hidden flaw in training on environments that follow a fixed sequence of levels. These results emphasize the importance of both training and evaluating with diverse environment distributions.

\subsection{500 Level Generalization}

Due to the high computational cost, it is impractical to regularly run the experiments described in \Cref{sec:level_gen}. To benchmark generalization, we recommend training on 500 levels from each environment and testing on held out levels, as in \cite{coinrun}. We choose this training set size to be near the region where generalization begins to take effect, as seen in \Cref{fig:gen_all}. At test time, we measure agents' zero-shot performance averaged over unseen levels. When evaluating generalization, we do not explicitly restrict the duration of training, though in practice we still train for 200M timesteps.

Baseline results are shown in \Cref{fig:gen500}. We see a high amount of overfitting in most environments. In some environments, the generalization gap is relatively small only because both training and test performance are poor, as discussed in \Cref{sec:level_gen}. In any case, we expect to see significant improvement on test performance as we develop agents more capable of generalization.

\section{Scaling Model Size} \label{sec:network_size}

We now investigate how scaling model size impacts both sample efficiency and generalization in RL. We conduct these experiments to demonstrate the usefulness of Procgen Benchmark metrics, and because this is a compelling topic in its own right. We follow the experimental protocols described in \Cref{sec:protocols}, evaluating the performance of 4 different models on both sample efficiency and generalization.

\begin{figure*}
\centering
\begin{subfigure}{0.475 \textwidth}
\includegraphics[width=\textwidth]{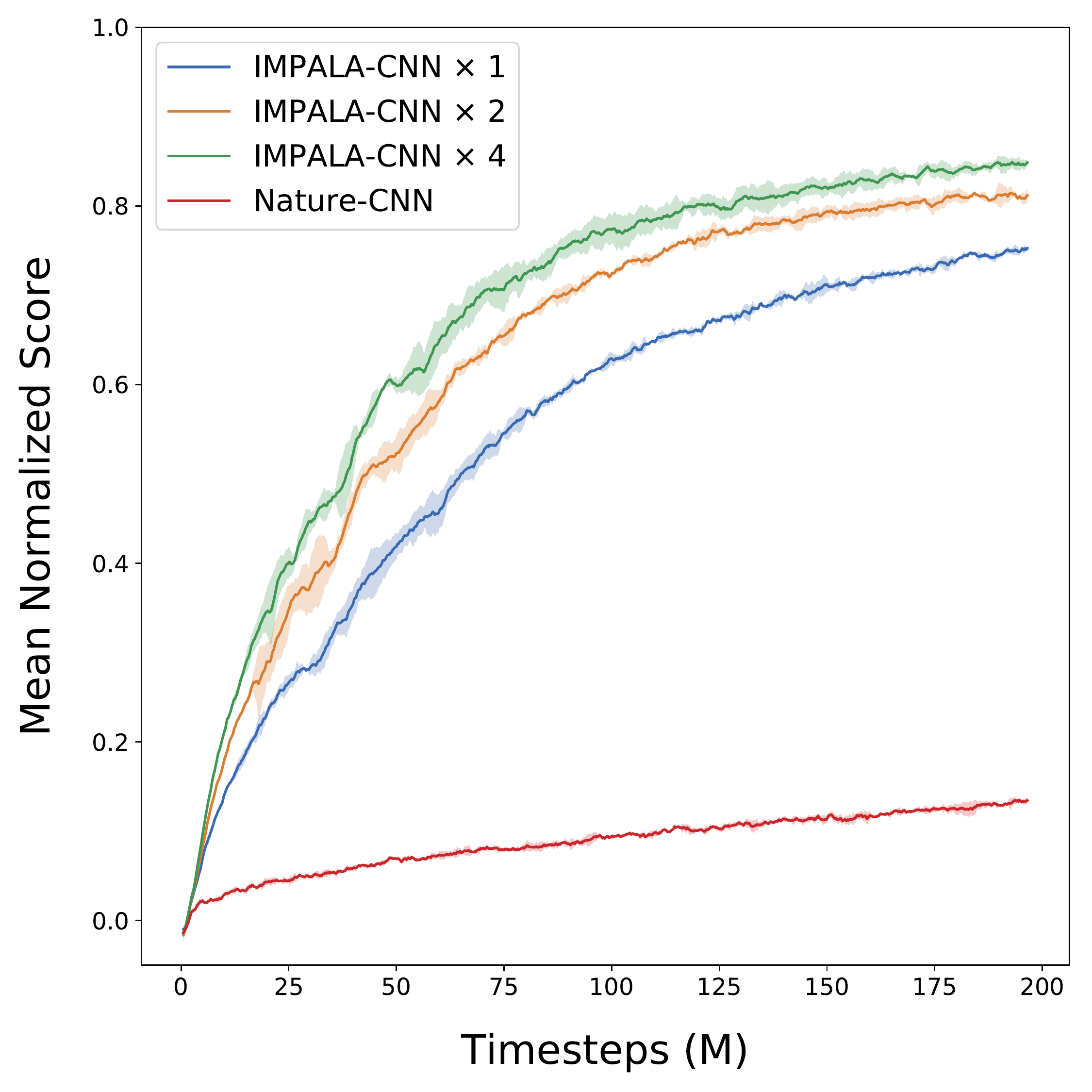}
\end{subfigure}
\hspace*{\fill}
\begin{subfigure}{0.475 \textwidth}
\includegraphics[width=\textwidth]{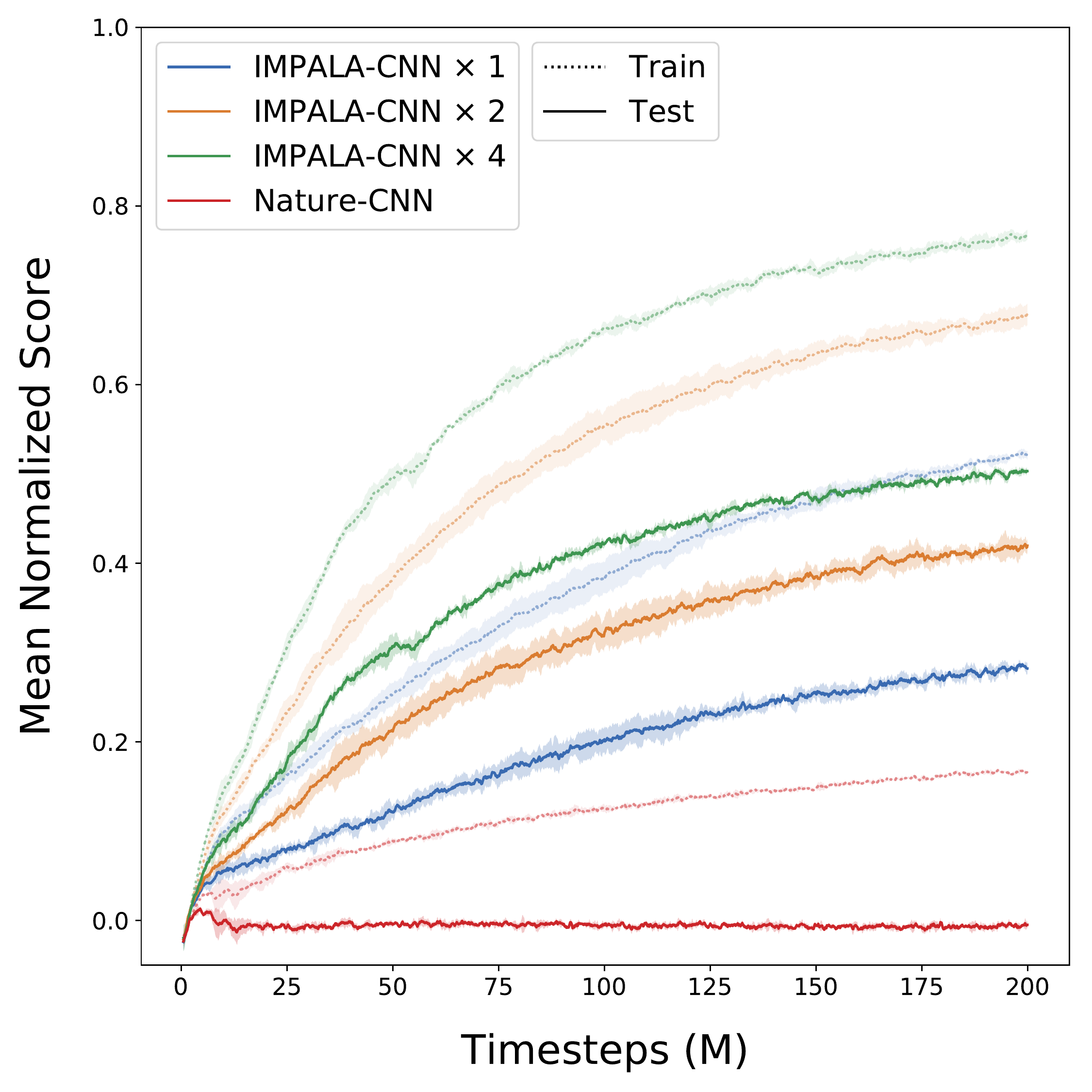}
\end{subfigure}
\caption{Performance of different model sizes, measuring both sample efficiency (left) and generalization (right). The mean and standard deviation is shown across 3 seeds.}
\label{fig:arch_cmp}
\end{figure*}

The first 3 models use the convolutional architecture found in IMPALA \citep{impala} with the number of convolutional channels at each layer scaled by $1$, $2$ or $4$. Note that scaling the number of channels by $k$ results in scaling the total parameter count by approximately $k^2$. The final model uses the smaller and more basic convolutional architecture found in \cite{dqn_2}, which we call Nature-CNN. We include this architecture as it is often used to train agents in the ALE.

We train the Nature-CNN model with the same learning rate as the smallest IMPALA model. When we scale the number of IMPALA channels by $k$, we also scale the learning rate by $\frac{1}{\sqrt k}$ to match the scaling of the weights, initialized with the method from \cite{glorot}. The learning rate is the only hyperparameter we vary between architectures. We performed sweeps over other hyperparameters, including the batch size and the number of epochs per rollout, and we found no other obvious gains.

Results are shown in \Cref{fig:arch_cmp}. We find that larger architectures significantly improve both sample efficiency and generalization. It is notable that the small Nature-CNN model almost completely fails to train. These results align with the results from \cite{coinrun}, and we now establish that this trend holds across many diverse environments. Although larger models offer fairly consistent improvements, we note that some environments benefit from the larger models to a greater extent. See \Cref{appendix:arch_cmp} for detailed training curves from each environment. 

\section{Comparing Algorithms} \label{sec:rainbow}

\begin{figure*}
\centering
\includegraphics[width=.9 \textwidth]{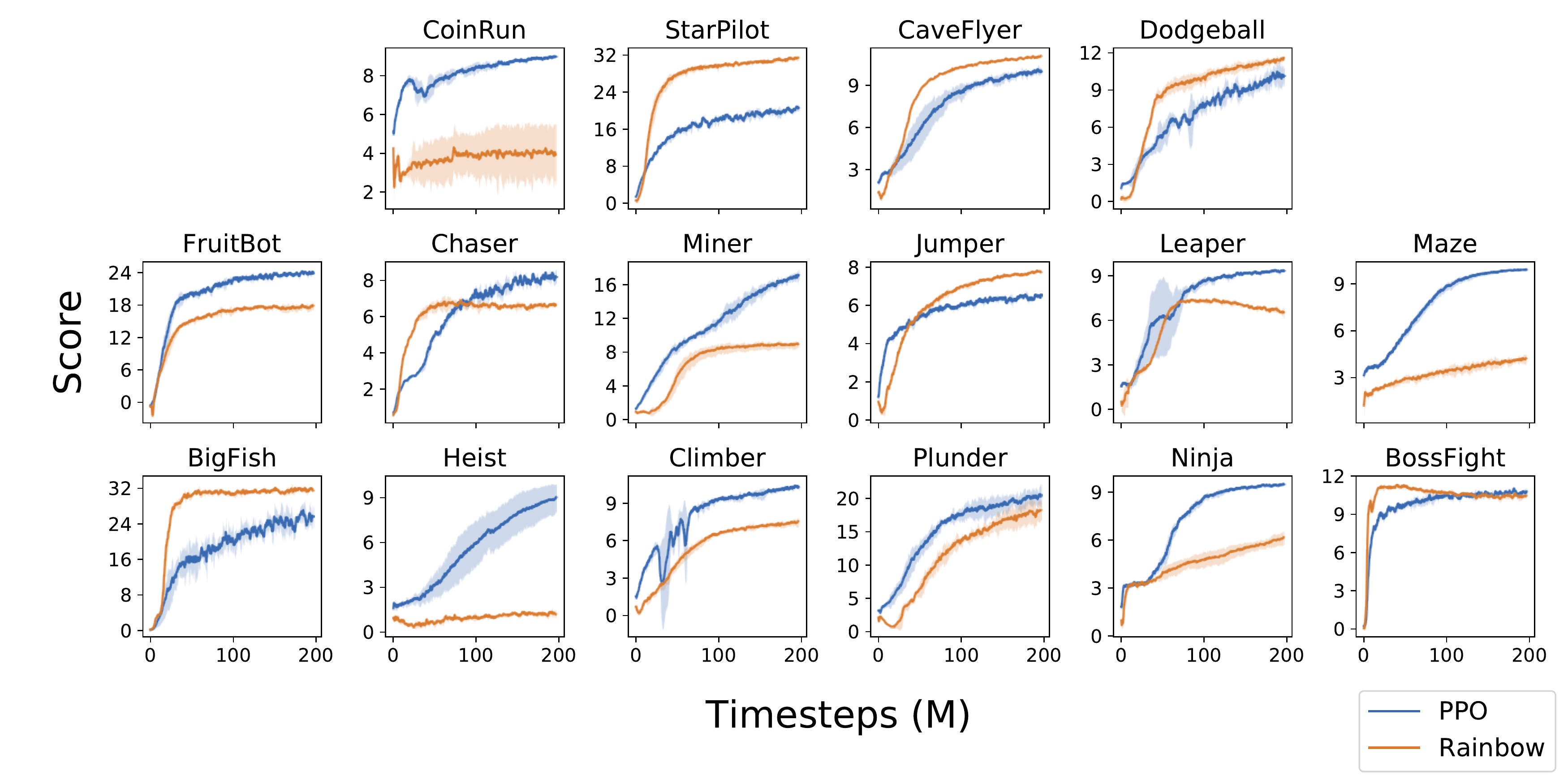}
\caption{A comparison between Rainbow and PPO. In both cases, we train and test on the full distribution of levels from each environment. The mean and standard deviation is shown across 3 seeds.}
\label{fig:rainbow}
\end{figure*}

We next compare our implementation of PPO to our implementation of Rainbow \citep{rainbow} on Procgen Benchmark. We evaluate sample efficiency, training and testing on the full distribution of levels in each environment. As with PPO, we train Rainbow agents using the IMPALA convolutional architecture, collecting experience from 64 parallel environment copies into a single replay buffer. We first experimented with the default Rainbow hyperparameters (with an appropriate choice for distributional min/max values), but we found that agents struggled to learn any non-trivial behaviour. We hypothesize that the diversity of our environments can lead to high variance gradients that promote instability. We therefore reduced gradient variance by running the algorithm on 8 parallel workers, using shared model parameters and averaging gradients between workers. This greatly improved performance.

To improve wall-clock time for Rainbow, we also increased the batch size and decreased the update frequency each by a factor of 16, while increasing the learning rate by a factor of 4. While this change significantly reduced wall-clock training time, it did not adversely impact performance. We confirmed that the new learning rate was roughly optimal by sweeping over nearby learning rates. See \Cref{appendix:hyperparameters} for a full list of Rainbow hyperparameters.

Results are shown in \Cref{fig:rainbow}. PPO performs much more consistently across the benchmark, though Rainbow offers a significant improvement in several environments. We're not presently able to diagnose the instability that leads to Rainbow's low performance in some environments, though we consider this an interesting avenue for further research.

\section{Related Work}

Many recent RL benchmarks grapple with generalization in different ways. The Sonic benchmark \citep{gottalearnfast} was designed to measure generalization in RL by separating levels of the $\textit{Sonic the Hedgehog}^\text{TM}$ video game into training and test sets. However, RL agents struggled to generalize from the few available training levels, and progress was hard to measure. The CoinRun environment \citep{coinrun} addressed this concern by procedurally generating large training and test sets to better measure generalization. CoinRun serves as the inaugural environment in Procgen Benchmark.

The General Video Game AI (GVG-AI) framework \citep{gvgai} has also encouraged the use of procedural generation in deep RL. Using 4 procedurally generated environments based on classic video games, \cite{illuminating_gen} measured generalization across different level distributions, finding that agents strongly overfit to their particular training set. Environments in Procgen Benchmark are designed in a similar spirit, with two of the environments (Miner and Leaper) drawing direct inspiration from this work.

The Obstacle Tower environment \citep{obstacle} attempts to measure generalization in vision, control, and planning using a 3D, 3rd person, procedurally generated environment. Success requires agents to solve both low-level control and high-level planning problems. While studying generalization in a single complex environment offers certain advantages, we opted to design many heterogeneous environments for Procgen Benchmark.

bsuite \citep{osband2019bsuite} is a set of simple environments designed to serve as ``an MNIST for reinforcement learning.'' Each environment targets a small number of core RL capabilities, including the core capability of generalization. bsuite includes a single environment that requires visual generalization in the form of an MNIST contextual bandit, whereas visual generalization is a primary source of difficulty across all Procgen environments.

Safety Gym \citep{safetygym} provides a suite of benchmark environments designed for studying safe exploration and constrained RL. While generalization is not an explicit focus of this benchmark, all Safety Gym environments perform extensive randomization to prevent agents from overfitting to specific environment layouts. In doing so, these environments enforce a need for generalization.

The Animal-AI Environment \citep{animalai} uses tasks inspired by the animal cognition literature to evaluate agent intelligence. Since these tests are not encountered during training, high performance depends on generalizing well from the specific training configurations. The use of a single unified environment makes the prospect of generalization significantly more plausible. 

Meta-World \citep{metaworld} proposes several meta-learning benchmarks, using up to 50 unique continuous control environments for training and testing. As with the Animal-AI Environment, the shared physics and mechanics between train and test environments gives rise to the plausible expectation of generalization, even when the details of the test task are novel.

\section{Conclusion}

Training agents capable of generalizing across environments remains one of the greatest challenges in reinforcement learning. We've designed Procgen Benchmark to help the community to contend with this challenge. The intrinsic diversity within level distributions makes this benchmark ideal for evaluating both generalization and sample efficiency in RL. We expect many insights gleaned from this benchmark to apply in more complex settings, and we look forward to leveraging these environments to design more capable and efficient algorithms.

\changeurlcolor{black}

\bibliography{leveraging_procgen}

\begin{thebibliography}{27}
\providecommand{\natexlab}[1]{#1}
\providecommand{\url}[1]{\texttt{#1}}
\expandafter\ifx\csname urlstyle\endcsname\relax
  \providecommand{\doi}[1]{doi: #1}\else
  \providecommand{\doi}{doi: \begingroup \urlstyle{rm}\Url}\fi

\bibitem[Achiam et~al.(2019)Achiam, Ray, and Amodei]{safetygym}
J.~Achiam, A.~Ray, and D.~Amodei.
\newblock Safety gym.
\newblock \url{https://openai.com/blog/safety-gym/}, 2019.

\bibitem[{Bellemare} et~al.(2013){Bellemare}, {Naddaf}, {Veness}, and
  {Bowling}]{ale}
M.~G. {Bellemare}, Y.~{Naddaf}, J.~{Veness}, and M.~{Bowling}.
\newblock The arcade learning environment: An evaluation platform for general
  agents.
\newblock \emph{Journal of Artificial Intelligence Research}, 47:\penalty0
  253--279, jun 2013.

\bibitem[Beyret et~al.(2019)Beyret, Hern'andez-Orallo, Cheke, Halina, Shanahan,
  and Crosby]{animalai}
B.~Beyret, J.~Hern'andez-Orallo, L.~Cheke, M.~Halina, M.~Shanahan, and
  M.~Crosby.
\newblock The animal-ai environment: Training and testing animal-like
  artificial cognition.
\newblock 2019.

\bibitem[Cobbe et~al.(2019)Cobbe, Klimov, Hesse, Kim, and Schulman]{coinrun}
K.~Cobbe, O.~Klimov, C.~Hesse, T.~Kim, and J.~Schulman.
\newblock Quantifying generalization in reinforcement learning.
\newblock In \emph{Proceedings of the 36th International Conference on Machine
  Learning, {ICML} 2019, 9-15 June 2019, Long Beach, California, {USA}}, pages
  1282--1289, 2019.
\newblock URL \url{http://proceedings.mlr.press/v97/cobbe19a.html}.

\bibitem[Espeholt et~al.(2018)Espeholt, Soyer, Munos, Simonyan, Mnih, Ward,
  Doron, Firoiu, Harley, Dunning, Legg, and Kavukcuoglu]{impala}
L.~Espeholt, H.~Soyer, R.~Munos, K.~Simonyan, V.~Mnih, T.~Ward, Y.~Doron,
  V.~Firoiu, T.~Harley, I.~Dunning, S.~Legg, and K.~Kavukcuoglu.
\newblock {IMPALA:} scalable distributed deep-rl with importance weighted
  actor-learner architectures.
\newblock \emph{CoRR}, abs/1802.01561, 2018.

\bibitem[Farebrother et~al.(2018)Farebrother, Machado, and Bowling]{gen_dqn}
J.~Farebrother, M.~C. Machado, and M.~Bowling.
\newblock Generalization and regularization in {DQN}.
\newblock \emph{CoRR}, abs/1810.00123, 2018.
\newblock URL \url{http://arxiv.org/abs/1810.00123}.

\bibitem[Glorot and Bengio(2010)]{glorot}
X.~Glorot and Y.~Bengio.
\newblock Understanding the difficulty of training deep feedforward neural
  networks.
\newblock In \emph{Proceedings of the thirteenth international conference on
  artificial intelligence and statistics}, pages 249--256, 2010.

\bibitem[Hessel et~al.(2018)Hessel, Modayil, Van~Hasselt, Schaul, Ostrovski,
  Dabney, Horgan, Piot, Azar, and Silver]{rainbow}
M.~Hessel, J.~Modayil, H.~Van~Hasselt, T.~Schaul, G.~Ostrovski, W.~Dabney,
  D.~Horgan, B.~Piot, M.~Azar, and D.~Silver.
\newblock Rainbow: Combining improvements in deep reinforcement learning.
\newblock In \emph{Thirty-Second AAAI Conference on Artificial Intelligence},
  2018.

\bibitem[Hochreiter and Schmidhuber(1997)]{lstm}
S.~Hochreiter and J.~Schmidhuber.
\newblock Long short-term memory.
\newblock \emph{Neural Computation}, 9\penalty0 (8):\penalty0 1735--1780, 1997.

\bibitem[Johnson et~al.(2010)Johnson, Yannakakis, and
  Togelius]{johnson2010cellular}
L.~Johnson, G.~N. Yannakakis, and J.~Togelius.
\newblock Cellular automata for real-time generation of infinite cave levels.
\newblock In \emph{Proceedings of the 2010 Workshop on Procedural Content
  Generation in Games}, page~10. ACM, 2010.

\bibitem[Juliani et~al.(2019)Juliani, Khalifa, Berges, Harper, Henry, Crespi,
  Togelius, and Lange]{obstacle}
A.~Juliani, A.~Khalifa, V.-P. Berges, J.~Harper, H.~Henry, A.~Crespi,
  J.~Togelius, and D.~Lange.
\newblock Obstacle tower: A generalization challenge in vision, control, and
  planning.
\newblock \emph{arXiv preprint arXiv:1902.01378}, 2019.

\bibitem[Justesen et~al.(2018)Justesen, Torrado, Bontrager, Khalifa, Togelius,
  and Risi]{illuminating_gen}
N.~Justesen, R.~R. Torrado, P.~Bontrager, A.~Khalifa, J.~Togelius, and S.~Risi.
\newblock Illuminating generalization in deep reinforcement learning through
  procedural level generation.
\newblock \emph{CoRR}, abs/1806.10729, 2018.

\bibitem[Kingma and Ba(2014)]{kingma2014adam}
D.~P. Kingma and J.~Ba.
\newblock Adam: A method for stochastic optimization.
\newblock \emph{arXiv preprint arXiv:1412.6980}, 2014.

\bibitem[Kruskal(1956)]{kruskal}
J.~B. Kruskal.
\newblock On the shortest spanning subtree of a graph and the traveling
  salesman problem.
\newblock In \emph{Proceedings of the American Mathematical Society, 7}, pages
  48--50, 1956.

\bibitem[Lee et~al.(2019)Lee, Lee, Shin, and Lee]{lee2019simple}
K.~Lee, K.~Lee, J.~Shin, and H.~Lee.
\newblock A simple randomization technique for generalization in deep
  reinforcement learning.
\newblock \emph{arXiv preprint arXiv:1910.05396}, 2019.

\bibitem[Machado et~al.(2018)Machado, Bellemare, Talvitie, Veness, Hausknecht,
  and Bowling]{machado18arcade}
M.~C. Machado, M.~G. Bellemare, E.~Talvitie, J.~Veness, M.~J. Hausknecht, and
  M.~Bowling.
\newblock Revisiting the arcade learning environment: Evaluation protocols and
  open problems for general agents.
\newblock \emph{Journal of Artificial Intelligence Research}, 61:\penalty0
  523--562, 2018.

\bibitem[Mnih et~al.(2015)Mnih, Kavukcuoglu, Silver, Rusu, Veness, Bellemare,
  Graves, Riedmiller, Fidjeland, Ostrovski, Petersen, Beattie, Sadik,
  Antonoglou, King, Kumaran, Wierstra, Legg, and Hassabis]{dqn_2}
V.~Mnih, K.~Kavukcuoglu, D.~Silver, A.~A. Rusu, J.~Veness, M.~G. Bellemare,
  A.~Graves, M.~A. Riedmiller, A.~Fidjeland, G.~Ostrovski, S.~Petersen,
  C.~Beattie, A.~Sadik, I.~Antonoglou, H.~King, D.~Kumaran, D.~Wierstra,
  S.~Legg, and D.~Hassabis.
\newblock Human-level control through deep reinforcement learning.
\newblock \emph{Nature}, 518\penalty0 (7540):\penalty0 529--533, 2015.

\bibitem[Nichol et~al.(2018)Nichol, Pfau, Hesse, Klimov, and
  Schulman]{gottalearnfast}
A.~Nichol, V.~Pfau, C.~Hesse, O.~Klimov, and J.~Schulman.
\newblock Gotta learn fast: {A} new benchmark for generalization in {RL}.
\newblock \emph{CoRR}, abs/1804.03720, 2018.
\newblock URL \url{http://arxiv.org/abs/1804.03720}.

\bibitem[Osband et~al.(2019)Osband, Doron, Hessel, Aslanides, Sezener, Saraiva,
  McKinney, Lattimore, {Sz}epesv{\'a}ri, Singh, Van~Roy, Sutton, Silver, and
  van Hasselt]{osband2019bsuite}
I.~Osband, Y.~Doron, M.~Hessel, J.~Aslanides, E.~Sezener, A.~Saraiva,
  K.~McKinney, T.~Lattimore, C.~{Sz}epesv{\'a}ri, S.~Singh, B.~Van~Roy,
  R.~Sutton, D.~Silver, and H.~van Hasselt.
\newblock Behaviour suite for reinforcement learning.
\newblock 2019.

\bibitem[Packer et~al.(2018)Packer, Gao, Kos, Kr{\"{a}}henb{\"{u}}hl, Koltun,
  and Song]{assess_gen_rl}
C.~Packer, K.~Gao, J.~Kos, P.~Kr{\"{a}}henb{\"{u}}hl, V.~Koltun, and D.~Song.
\newblock Assessing generalization in deep reinforcement learning.
\newblock \emph{CoRR}, abs/1810.12282, 2018.

\bibitem[Perez-Liebana et~al.(2018)Perez-Liebana, Liu, Khalifa, Gaina,
  Togelius, and Lucas]{gvgai}
D.~Perez-Liebana, J.~Liu, A.~Khalifa, R.~D. Gaina, J.~Togelius, and S.~M.
  Lucas.
\newblock General video game ai: a multi-track framework for evaluating agents,
  games and content generation algorithms.
\newblock \emph{arXiv preprint arXiv:1802.10363}, 2018.

\bibitem[Pfau et~al.(2018)Pfau, Nichol, Hesse, Schiavo, Schulman, and
  Klimov]{gymretro}
V.~Pfau, A.~Nichol, C.~Hesse, L.~Schiavo, J.~Schulman, and O.~Klimov.
\newblock Gym retro.
\newblock \url{https://openai.com/blog/gym-retro/}, 2018.

\bibitem[Schulman et~al.(2017)Schulman, Wolski, Dhariwal, Radford, and
  Klimov]{ppo}
J.~Schulman, F.~Wolski, P.~Dhariwal, A.~Radford, and O.~Klimov.
\newblock Proximal policy optimization algorithms.
\newblock \emph{CoRR}, abs/1707.06347, 2017.

\bibitem[Yu et~al.(2019)Yu, Quillen, He, Julian, Hausman, Finn, and
  Levine]{metaworld}
T.~Yu, D.~Quillen, Z.~He, R.~Julian, K.~Hausman, C.~Finn, and S.~Levine.
\newblock Meta-world: A benchmark and evaluation for multi-task and
  meta-reinforcement learning, 2019.
\newblock URL \url{https://github.com/rlworkgroup/metaworld}.

\bibitem[Zhang et~al.(2018{\natexlab{a}})Zhang, Ballas, and
  Pineau]{dissect_overfitting}
A.~Zhang, N.~Ballas, and J.~Pineau.
\newblock A dissection of overfitting and generalization in continuous
  reinforcement learning.
\newblock \emph{CoRR}, abs/1806.07937, 2018{\natexlab{a}}.

\bibitem[Zhang et~al.(2018{\natexlab{b}})Zhang, Wu, and Pineau]{natural_ale}
A.~Zhang, Y.~Wu, and J.~Pineau.
\newblock Natural environment benchmarks for reinforcement learning.
\newblock \emph{arXiv preprint arXiv:1811.06032}, 2018{\natexlab{b}}.

\bibitem[Zhang et~al.(2018{\natexlab{c}})Zhang, Vinyals, Munos, and
  Bengio]{study_overfitting}
C.~Zhang, O.~Vinyals, R.~Munos, and S.~Bengio.
\newblock A study on overfitting in deep reinforcement learning.
\newblock \emph{CoRR}, abs/1804.06893, 2018{\natexlab{c}}.
\newblock URL \url{http://arxiv.org/abs/1804.06893}.

\end{thebibliography}

\changeurlcolor{blue}

\appendix

\clearpage

\section{Environment Descriptions} \label{appendix:env_disc}

In all environments, procedural generation controls the selection of game assets and backgrounds, though some environments include a more diverse pool of assets and backgrounds than others. When procedural generation must place entities, it generally samples from the uniform distribution over valid locations, occasionally subject to game-specific constraints. Several environments use cellular automata \citep{johnson2010cellular} to generate diverse level layouts.

\subsection{CoinRun}

A simple platformer. The goal is to collect the coin at the far right of the level, and the player spawns on the far left. The player must dodge stationary saw obstacles, enemies that pace back and forth, and chasms that lead to death. Note that while the previously released version of CoinRun painted velocity information directly onto observations, the current version does not. This makes the environment significantly more difficult.

Procedural generation controls the number of platform sections, their corresponding types, the location of crates, and the location and types of obstacles.

\subsection{StarPilot}

A simple side scrolling shooter game. All enemies fire projectiles that directly target the player, so an inability to dodge quickly leads to the player's demise. There are fast and slow enemies, stationary turrets with high health, clouds which obscure player vision, and impassable meteors.

Procedural generation controls the spawn timing of all enemies and obstacles, along with their corresponding types.

\subsection{CaveFlyer}

The player controlling a starship must navigate a network of caves to reach the goal (a friendly starship). Player movement mimics the Atari game “Asteroids”: the ship can rotate and travel forward or backward along the current axis. The majority of the reward comes from successfully reaching the goal, though additional reward can be collected by destroying target objects along the way with the ship's lasers. There are stationary and moving lethal obstacles throughout the level.

Procedural generation controls the level layout via cellular automata, as well as the configuration of all enemies, targets, obstacles, and the goal.

\subsection{Dodgeball}

Loosely inspired by the Atari game “Berzerk”. The player spawns in a room with walls and enemies. Touching a wall loses the game and ends the episode. The player moves relatively slowly and can navigate throughout the room. There are enemies which also move slowly and which will occasionally throw balls at the player. The player can also throw balls, but only in the direction they are facing. If all enemies are hit, the player can move to the unlocked platform and earn a significant level completion bonus.

Procedural generation controls the level layout by recursively generating room-like structures. It also controls the quantity and configuration of enemies.

\subsection{FruitBot}

A scrolling game where the player controls a robot that must navigate between gaps in walls and collect fruit along the way. The player receives a positive reward for collecting a piece of fruit, and a larger negative reward for mistakenly collecting a non-fruit object. On expectation, half of the spawned objects are fruit (positive reward) and half are non-fruit (negative reward). The player receives a large reward if they reach the end of the level. Occasionally the player must use a key to unlock gates which block the way.

Procedural generation controls the level layout by sequentially generating barriers with randomly-sized gaps. It also controls the quantity and configuration of fruit and non-fruit objects, as well as the placement of gates.

\subsection{Chaser}

Inspired by the Atari game “MsPacman”. The player must collect all the green orbs in the level. 3 large stars spawn that will make enemies vulnerable for a short time when collected. A collision with an enemy that isn’t vulnerable results in the player’s death. When a vulnerable enemy is eaten, an egg spawns somewhere on the map that will hatch into a new enemy after a short time, keeping the total number of enemies constant. The player receives a small reward for collecting each orb and a large reward for completing the level.

Procedural generation controls the level layout by generating mazes using Kruskal’s algorithm \citep{kruskal}, and then removing walls until no dead-ends remain. The large stars are constrained to spawn in different quadrants. Initial enemy spawn locations are randomly selected.

\subsection{Miner}

Inspired by the classic game “BoulderDash”. The player, a robot, can dig through dirt to move throughout the world. The world has gravity, and dirt supports boulders and diamonds. Boulders and diamonds will fall through free space and roll off each other. If a boulder or a diamond falls on the player, the game is over. The goal is to collect all the diamonds in the level and then proceed through the exit. The player receives a small reward for collecting a diamond and a larger reward for completing the level.

Procedural generation controls the position of all boulders, diamonds, and the exit. No objects may spawn adjacent to the player. An approximately fixed quantity of boulders and diamonds spawn in each level.

\subsection{Jumper}

A platformer with an open world layout. The player, a bunny, must navigate through the world to find the carrot. It might be necessary to ascend or descend the level to do so. The player is capable of “double jumping”, allowing it to navigate tricky layouts and reach high platforms. There are spike obstacles which will destroy the player on contact. The screen includes a compass which displays direction and distance to the carrot. The only reward in the game comes from collect the carrot, at which point the episode ends.

Procedural generation controls the level layout via cellular automata, which is seeded with a maze-like structure. Long flat vertical edges are intentionally perturbed to avoid unsolvable levels, as the player can take advantage of irregular ledges on vertical walls. Obstacles cannot spawn adjacent to each other, as this could create impassable configurations.

\subsection{Leaper}

Inspired by the classic game “Frogger”. The player must cross several lanes to reach the finish line and earn a reward. The first group of lanes contains cars which must be avoided. The second group of lanes contains logs on a river. The player must hop from log to log to cross the river. If the player falls in the river, the episode ends.

Procedural generation controls the number of lanes of both roads and water, with these choices being positively correlated. It also controls the spawn timing of all logs and cars.

\subsection{Maze}

The player, a mouse, must navigate a maze to find the sole piece of cheese and earn a reward. The player may move up, down, left or right to navigate the maze.

Procedural generation controls the level layout by generating mazes using Kruskal’s algorithm \citep{kruskal}, uniformly ranging in size from 3x3 to 25x25.

\subsection{BigFish}

The player starts as a small fish and becomes bigger by eating other fish. The player may only eat fish smaller than itself, as determined solely by width. If the player comes in contact with a larger fish, the player is eaten and the episode ends. The player receives a small reward for eating a smaller fish and a large reward for becoming bigger than all other fish, at which point the episode ends. 

Procedural generation controls the spawn timing and position of all fish.

\subsection{Heist}

The player must steal the gem hidden behind a network of locks. Each lock comes in one of three colors, and the necessary keys to open these locks are scattered throughout the level. The level layout takes the form of a maze. Once the player collects a key of a certain color, the player may open the lock of that color. All keys in the player's possession are shown in the top right corner of the screen.

Procedural generation controls the level layout by generating mazes using Kruskal’s algorithm \citep{kruskal}. Locks and keys are randomly placed, subject to solvability constraints.

\subsection{Climber}

A simple platformer. The player must climb a sequence of platforms, collecting stars along the way. A small reward is given for collecting a star, and a larger reward is given for collecting all stars in a level. If all stars are collected, the episode ends. There are lethal flying monsters scattered throughout the level.

Procedural generation controls the level layout by sequentially generating reachable platforms. Enemies and stars spawn near each platform with fixed probabilities, except when spawning an enemy would lead to an unsolvable configuration. The final platform always contains a star.

\subsection{Plunder}

The player must destroy enemy pirate ships by firing cannonballs from its own ship at the bottom of the screen. An on-screen timer slowly counts down. If this timer runs out, the episode ends. Whenever the player fires, the timer skips forward a few steps, encouraging the player to conserve ammunition. The player should also avoid hitting friendly ships. The player receives a positive reward for hitting an enemy ship and a large timer penalty for hitting a friendly ship. A target in the bottom left corner identifies the color of the enemy ships to target. Wooden obstacles capable of blocking the player's line of sight may exist.

Procedural generation controls the selection of friendly and enemy ship types, as well as the spawn times and positions of all non-player ships. It also controls the placement of wooden obstacles.

\subsection{Ninja}

A simple platformer. The player, a ninja, must jump across narrow ledges while avoiding bomb obstacles. The player can toss throwing stars at several angles in order to clear bombs, if necessary. The player's jump can be charged over several timesteps to increase its effect. The player receives a reward for collecting the mushroom at the end of the level, at which point the episode terminates.

Procedural generation controls the level layout by sequentially generating reachable platforms, with the possibility of superfluous platform generation. Bombs are occasionally randomly placed near platforms.

\subsection{Bossfight}

The player controls a small starship and must destroy a much bigger boss starship. The boss randomly selects from a set of possible attacks when engaging the player. The player must dodge the incoming projectiles or be destroyed. The player can also use randomly scattered meteors for cover. After a set timeout, the boss becomes vulnerable and its shields go down. At this point, the players projectile attacks will damage the boss. Once the boss receives a certain amount of damage, the player receives a reward, and the boss re-raises its shields. If the player damages the boss several times in this way, the boss is destroyed, the player receives a large reward, and the episode ends.

Procedural generation controls certain game constants, including the boss health and the number of rounds in a level. It also selects the configuration of meteors in the level, and the attack pattern sequence the boss will follow.

\newpage

\section{Core Capabilities in RL}

To better understand the strengths and limitations of current RL algorithms, it is valuable to have environments which isolate critical axes of performance. \cite{osband2019bsuite} recently proposed seven core RL capabilities to profile with environments in bsuite. We focus our attention on three of these core capabilities: generalization, exploration, and memory. Among these, Procgen Benchmark contributes most directly to the evaluation of generalization, as we have already discussed at length. In this section, we describe how Procgen environments can also shed light on the core capabilities of exploration and memory.

\subsection{Evaluating Exploration} \label{appendix:exploration}

The trade off between exploration and exploitation has long been recognized as one of the principal challenges in reinforcement learning. Although exploration plays some role in every environment, the difficulty of the exploration problem can vary drastically. In many environments, the ability to adequately explore becomes an overwhelming bottleneck in agents' training. With Procgen environments, we strive to be deliberate in our consideration of exploration.

The generalization curves in \Cref{fig:gen_all} show that training performance often increases with the size of the training set. This reveals an interesting phenomenon: exploration can become less of a bottleneck in the presence of greater diversity. On the other hand, when the training set is restricted and diversity is removed, an otherwise tractable environment can become intractable due to exploration. By taking this to the extreme and restricting training to a single high difficulty level, 8 of the Procgen environments can be made into highly challenging exploration tasks. In doing so, these environments come to resemble traditional hard exploration environments, like the infamous Atari game Montezuma's Revenge. We note that generalization is not measured in this setting; the focus is solely on the agent's ability to explore.

The 8 environments that specifically support the evaluation of exploration are CoinRun, CaveFlyer, Leaper, Jumper, Maze, Heist, Climber, and Ninja. For each environment, we handpick a level seed that presents a significant exploration challenge. Instruction for training on these specific seeds can be found at \href{https://github.com/openai/train-procgen}{https://github.com/openai/train-procgen}. On these levels, a random agent is extraordinarily unlikely to encounter any reward. For this reason, our baseline PPO implementation completely fails to train, achieving a mean return of 0 in all environments after 200M timesteps of training.

\subsection{Evaluating Memory} \label{appendix:memory}

The extent to which agents must attend to the past varies greatly by environment. In environments in the ALE, memory beyond a small frame stack is not generally required to achieve optimal performance. In more general settings and in more complex environments, we expect memory to become increasingly relevant.

By default, Procgen environments require little to no use of memory, and non-recurrent policies achieve approximately the same level of performance as recurrent policies. We designed environments in this way to better isolate the challenges in RL. However, 6 of the 16 Procgen environments support variants that do require memory. These variants remove linearity constraints from level generation and increase the impact of partial observability. By introducing a dependence on memory, these environments become dramatically more difficult.

The 6 environments that specifically support the evaluation of memory are CoinRun, CaveFlyer, Dodgeball, Miner, Jumper, Maze, and Heist. In this setting, we modify the environments as follows. In all environments we increase the world size. In Caveflyer and Jumper, we remove logic in level generation that prunes away paths which do not lead to the goal. In Dodgeball, Miner, Maze, and Heist, we make the environments partially observable by restricting observations to a small patch of space surrounding the agent. We note that Caveflyer and Jumper were already partially observable. With these changes, agents can reliably solve levels only by utilizing memory. Instructions for training environments in memory mode can be found at \href{https://github.com/openai/train-procgen}{https://github.com/openai/train-procgen}.

\newpage

\section{Normalization Constants} \label{appendix:norm_constants}

$R_{min}$ is computed by training a policy with masked out observations. This demonstrates what score is trivially achievable in each environment. $R_{max}$ is computed in several different ways.

For CoinRun, Dodgeball, Miner, Jumper, Leaper, Maze, BigFish, Heist, Plunder, Ninja, and Bossfight, the maximal theoretical and practical reward is trivial to compute.

For CaveFlyer, Chaser, and Climber, we empirically determine $R_{max}$ by generating many levels and computing the average max achievable reward.

For StarPilot and FruitBot, the max practical reward is not obvious, even though it is easy to establish a theoretical bound. We choose to define $R_{max}$ in these environments as the score PPO achieves after 8 billion timesteps when trained at an 8x larger batch size than our default hyperparameters. On observing these policies, we find them very close to optimal.

\begin{center}
 \begin{tabular}{||c c c c c||} 
 \multicolumn{1}{c}{} & \multicolumn{2}{c}{Hard} & \multicolumn{2}{c}{Easy} \\
 \hline
 Environment & $R_{min}$ & $R_{max}$ & $R_{min}$ & $R_{max}$ \\
 \hline\hline
 CoinRun & 5 & 10 & 5 & 10 \\ 
 StarPilot & 1.5 & 35 & 2.5 & 64\\ 
 CaveFlyer & 2 & 13.4 & 3.5 & 12 \\ 
 Dodgeball & 1.5 & 19 & 1.5 & 19 \\ 
 FruitBot & -.5 & 27.2 & -1.5 & 32.4 \\ 
 Chaser & .5 & 14.2 & .5 & 13 \\ 
 Miner & 1.5 & 20 & 1.5 & 13 \\ 
 Jumper & 1 & 10 & 3 & 10 \\ 
 Leaper & 1.5 & 10 & 3 & 10 \\ 
 Maze & 4 & 10 & 5 & 10 \\ 
 BigFish & 0 & 40 & 1 & 40 \\ 
 Heist & 2 & 10 & 3.5 & 10 \\ 
 Climber & 1 & 12.6 & 2 & 12.6 \\ 
 Plunder & 3 & 30 & 4.5 & 30 \\ 
 Ninja & 2 & 10 & 3.5 & 10 \\ 
 BossFight & .5 & 13 & .5 & 13 \\ 
 \hline
 \end{tabular}
\end{center}

\onecolumn
\clearpage
\section{Hyperparameters} \label{appendix:hyperparameters}

We use the Adam optimizer \citep{kingma2014adam} in all experiments.

\begin{table}[h]
\caption{PPO Hyperparameters}
\label{ppo-table}
\vskip 0.15in
\begin{center}
\begin{small}
\begin{sc}
\begin{tabular}{lcc}
\toprule
Env. Distribution Mode & Hard & Easy \\
\midrule
$\gamma$ & .999 & .999 \\ 
$\lambda$ & .95 & .95 \\
\# timesteps per rollout & 256 & 256 \\
Epochs per rollout & 3 & 3 \\
\# minibatches per epoch & 8 & 8 \\ 
Entropy bonus ($k_H$) & .01 & .01 \\
PPO clip range & .2 & .2 \\
Reward Normalization? & Yes & Yes \\
Learning rate & \num{5e-4} & \num{5e-4} \\
\# workers & 4 & 1 \\
\# environments per worker & 64 & 64 \\ 
Total timesteps & 200M & 25M \\
LSTM? & No & No \\
Frame Stack? & No & No \\
\bottomrule
\end{tabular}
\end{sc}
\end{small}
\end{center}
\vskip -0.1in
\end{table}

\begin{table}[h]
\caption{Rainbow Hyperparameters}
\label{rainbow-table}
\vskip 0.15in
\begin{center}
\begin{small}
\begin{sc}
\begin{tabular}{lc}
\toprule
Env. Distribution Mode: & Hard \\
\midrule
$\gamma$ & .99 \\ 
Learning rate & \num{2.5e-4} \\
\# workers & 8 \\
\# environments per worker & 64 \\
\# env. steps per update per worker & 64 \\ 
Batch size per worker & 512 \\ 
Reward Clipping? & No \\
Distributional min/max values & [0, $R_{max}$]\footnotemark \\
Total timesteps & 200M \\
LSTM? & No \\
Frame Stack? & No \\
\bottomrule
\end{tabular}
\end{sc}
\end{small}
\end{center}
\vskip -0.1in
\end{table}

\footnotetext{In FruitBot, we change the distributional min value to -5.}

\newpage

\section{Test Performance for All Training Sets} \label{appendix:gen_test_curves}

\begin{figure}[h]
\centering
\includegraphics[width=.9\textwidth]{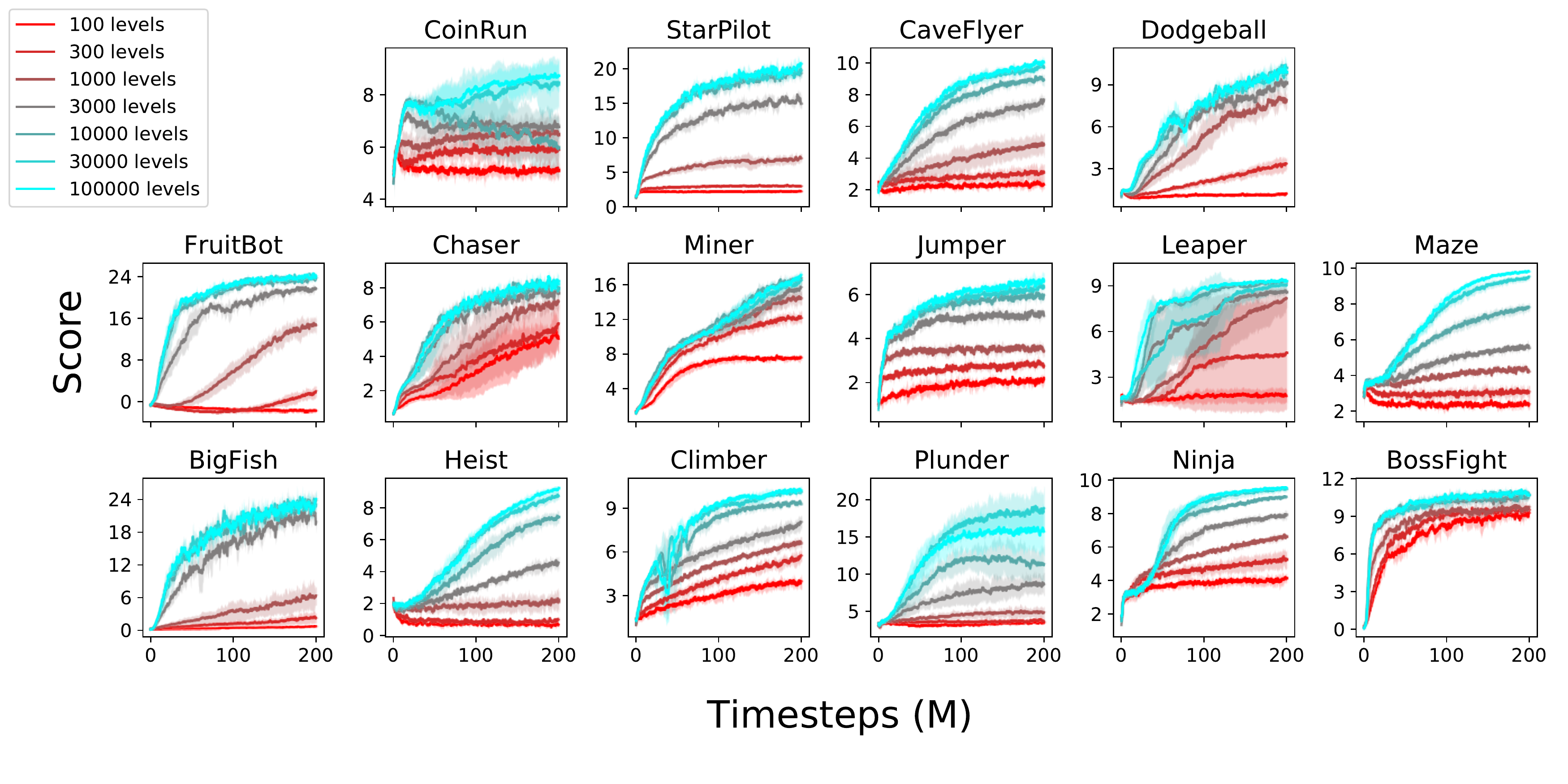}
\caption{Test performance of agents trained on different sets of levels. All agents are evaluated on the full distribution of levels from each environment.}
\end{figure}

\newpage

\section{Arcade Learning Environment Performance} \label{appendix:atari}

\begin{figure}[h]
\centering
\includegraphics[width=.9\textwidth]{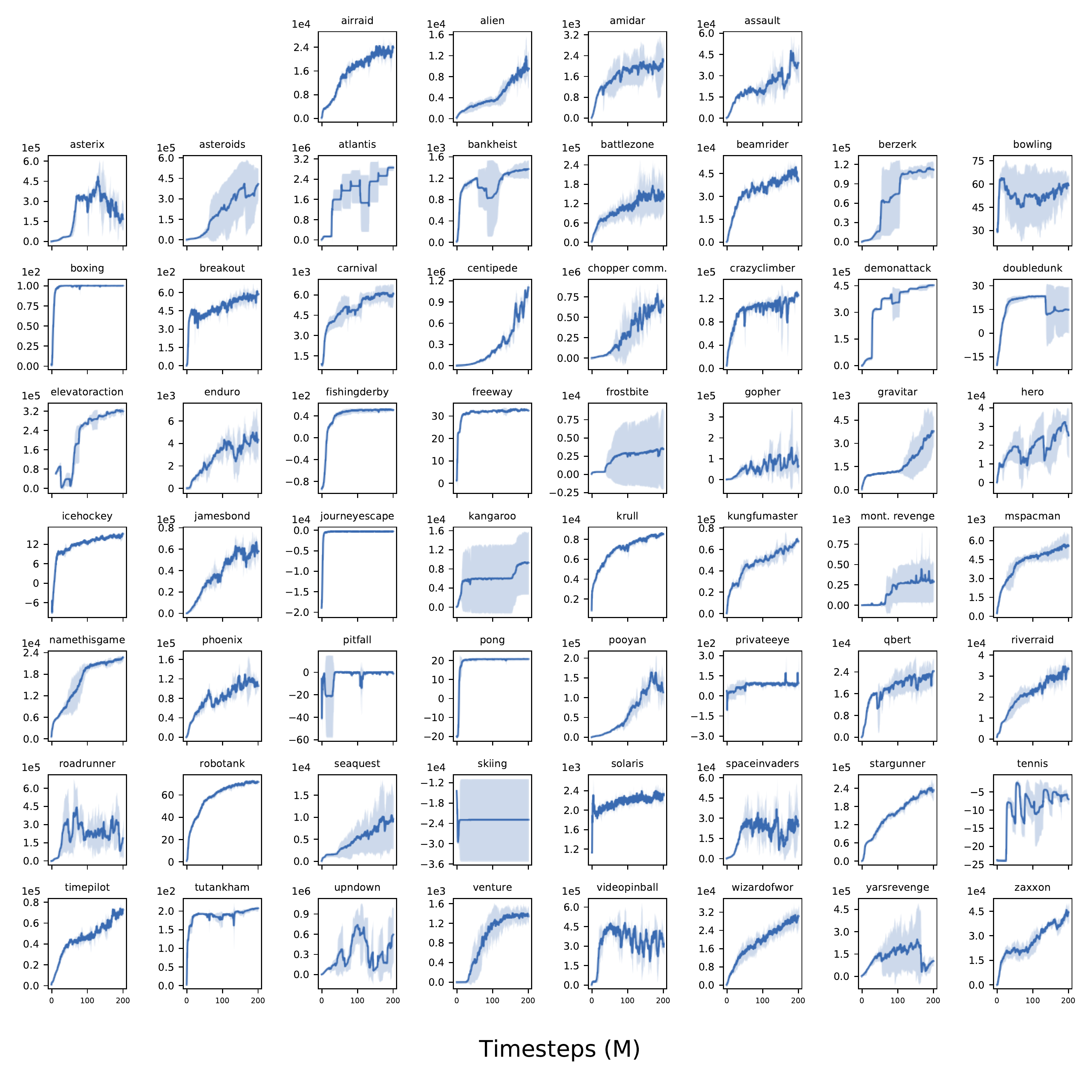}
\caption{Performance of our implementation of PPO on the ALE.}
\end{figure}

In these ALE experiments, we use a frame stack of 4 and we do not use sticky actions \citep{machado18arcade}. Note that although we render Procgen environments at 64x64 pixels, they can easily be rendered at 84x84 pixels to match the ALE standard, if desired.

\newpage

\section{Training Curves by Architecture} \label{appendix:arch_cmp}

\begin{figure}[h]
\centering
\includegraphics[width=.8\textwidth]{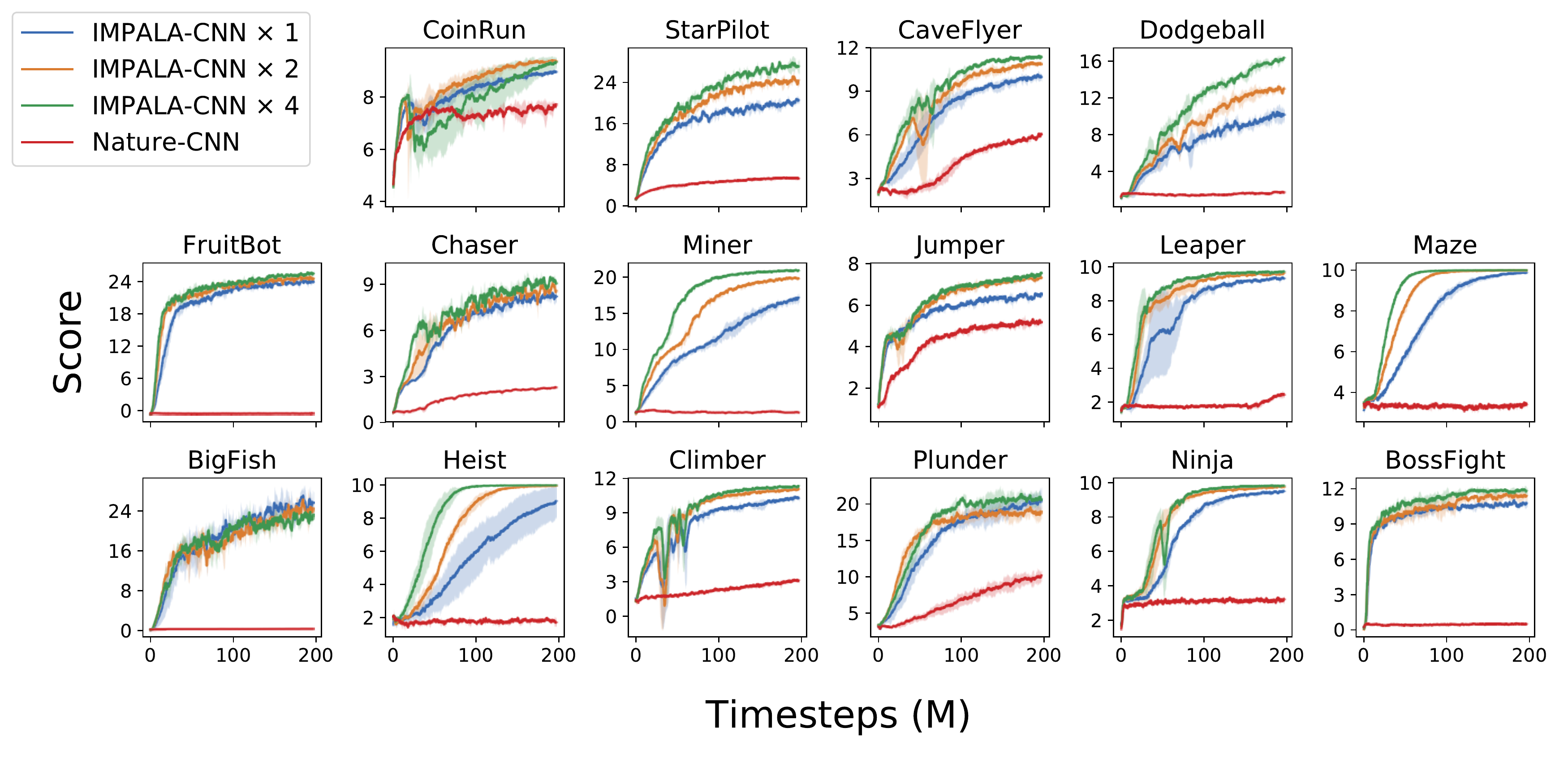}
\caption{Performance of agents using each different architecture in each environment, trained and evaluated on the full distribution of levels.}
\label{fig:atari}
\end{figure}

\begin{figure}[h]
\centering
\includegraphics[width=.8\textwidth]{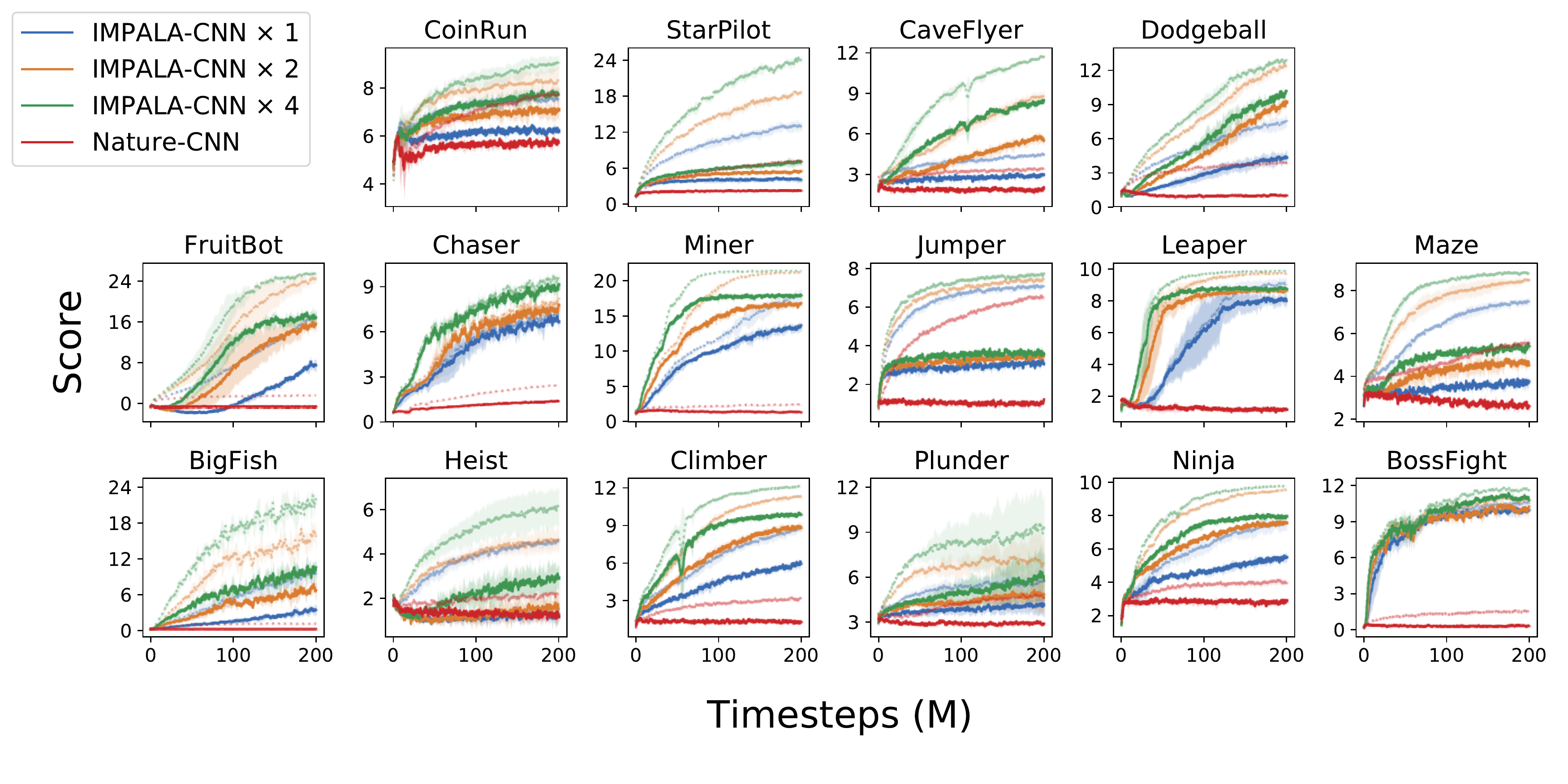}
\caption{Performance of agents using each different architecture in each environment, trained on 500 levels and evaluated on held out levels. Light dashed lines denote training curves and dark solid lines denote test curves.}
\end{figure}

\newpage

\section{Frame Stack vs. LSTM} \label{appendix:frame_stack}

\begin{figure}[h]
\includegraphics[width=\textwidth]{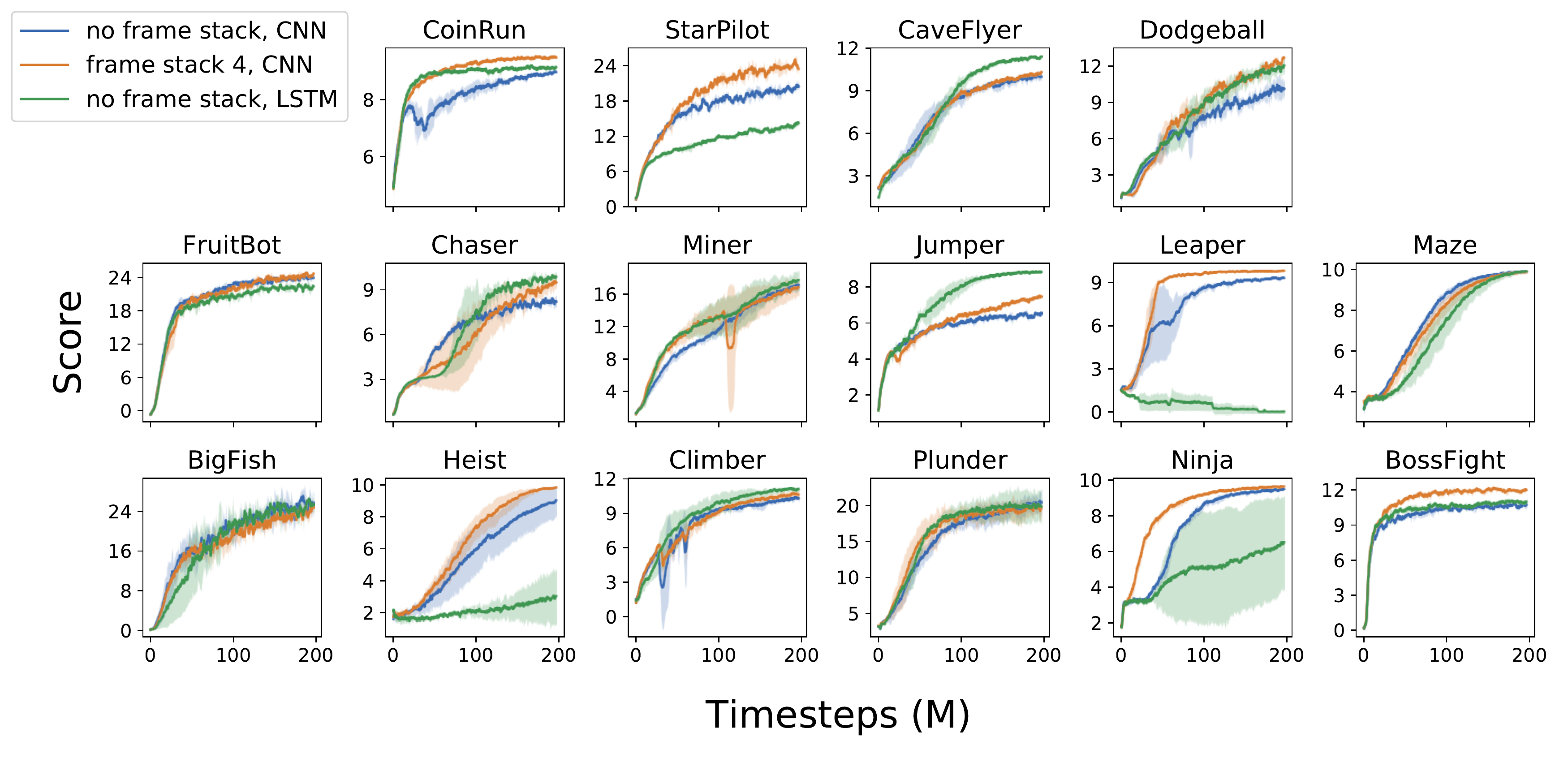}
\caption{Comparison of our baseline to agents that use either frame stack or a recurrent architecture.}
\end{figure}

For simplicity, our baseline experiments forgo the use of frame stack by default. This limits agents to processing information from only the single current frame. We compare this baseline to agents that use a frame stack of 4. We also compare both methods to agents using an LSTM \citep{lstm} on top of the convolutional network.

In general, we find these methods to be fairly comparable. In the Jumper environment, the LSTM agents outperform others, perhaps as the ability to perform non-trivial temporally extended navigation is helpful. In other environments like Leaper and Ninja, our LSTM baseline is notably unstable. In most environments, frame stack agents perform similarly to baseline agents, though in a few environments the difference is noticeable.

\newpage

\section{Easy Difficulty Baseline Results} \label{appendix:easy}

\begin{figure}[h]
\centering
\includegraphics[width=.8\textwidth]{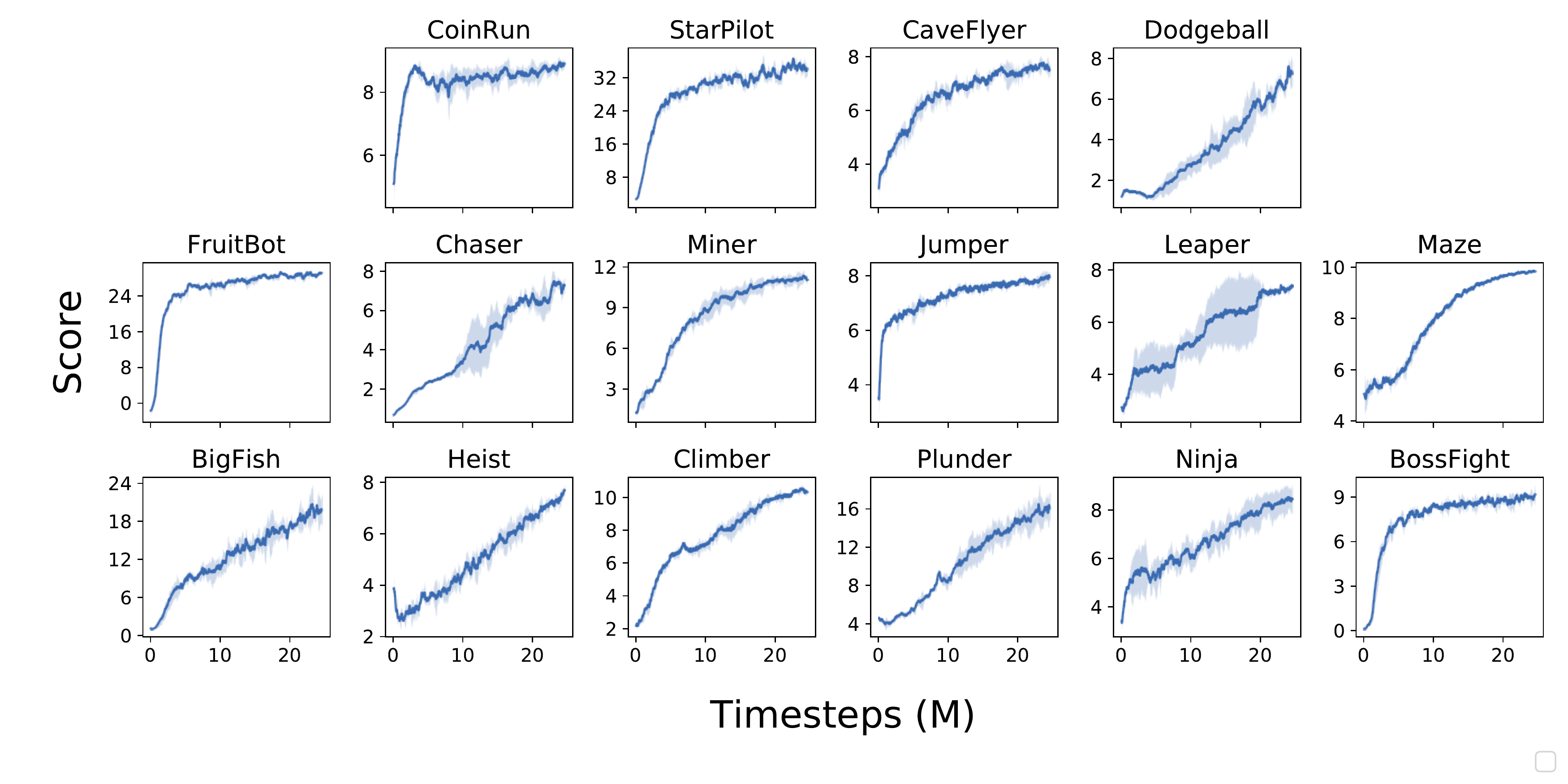}
\caption{Performance of agents on easy difficulty environments, trained and evaluated on the full distribution of levels.}
\end{figure}

\begin{figure}[h]
\centering
\includegraphics[width=.8\textwidth]{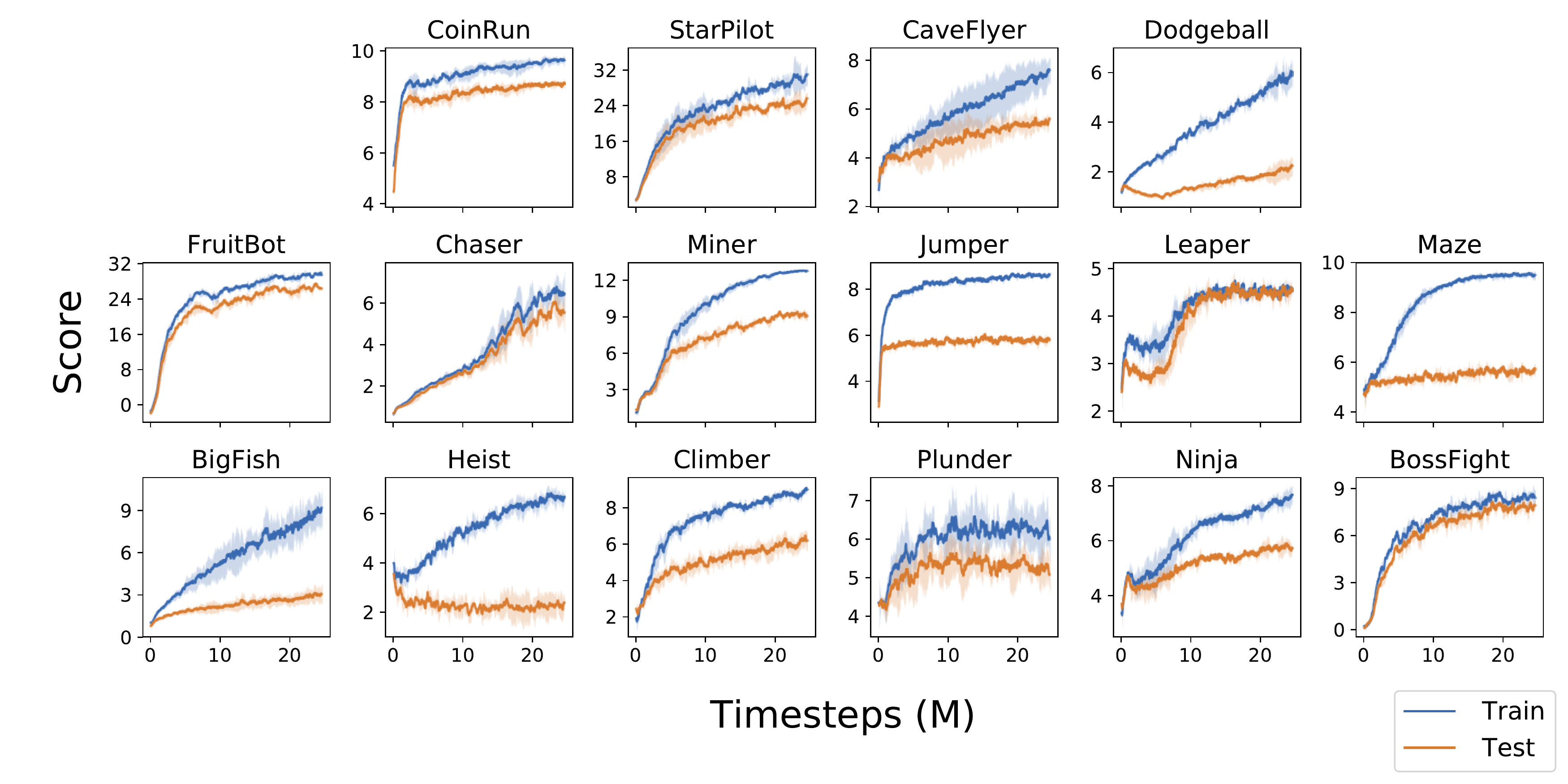}
\caption{Performance of agents on easy difficulty environments, trained on 200 levels and evaluated on the full distribution of levels.}
\end{figure}

\end{document}